\pdfoutput=1

\documentclass[11pt]{article}

\usepackage[]{ACL2023}

\usepackage{xcolor}
\usepackage{times}
\usepackage{latexsym}
\usepackage{booktabs}
\usepackage{enumitem}

\usepackage[T1]{fontenc}
\usepackage{graphicx}
\usepackage[utf8]{inputenc}

\usepackage{microtype}
\usepackage{amssymb}
\DeclareUnicodeCharacter{211D}{\mathbb{R}}
\usepackage{inconsolata}

\usepackage{soul}
\usepackage{multirow}
\usepackage{cleveref}

\title{A Comprehensive Evaluation of Large Language Models on \\ Benchmark Biomedical Text Processing Tasks}

 \author{Israt Jahan\textsuperscript{\textdagger, \textdollar}, Md Tahmid Rahman Laskar\thanks{\hspace{0.115cm} Corresponding Authors.}\textsuperscript{  \  ,\textdaggerdbl, \textdollar}, \\ \textbf{Chun Peng\textsuperscript{\textdagger}, Jimmy Xiangji Huang\footnotemark[1] \textsuperscript{ , \textdaggerdbl, \textdollar}} \\
 \textsuperscript{\textdagger}Department of Biology, York University\\ \textsuperscript{\textdaggerdbl}School of Information Technology, York University \\
  \textsuperscript{\textdollar}Information Retrieval and Knowledge Management Research Lab, York University \\
  Toronto, Ontario, Canada \\
  \texttt{\{israt18,tahmid20,cpeng,jhuang\}@yorku.ca}}

\begin{document}
\maketitle

\begin{abstract}
Recently, \textcolor{black}  {\textbf{L}arge \textbf{L}anguage \textbf{M}odel\textbf{s} (LLMs)} have demonstrated impressive capability to solve a wide range of tasks. However, despite their success across various tasks, no prior work has investigated their capability in the biomedical domain yet. To this end, this paper aims to evaluate the performance of LLMs on benchmark biomedical tasks. For this purpose, \textcolor{black} {a comprehensive evaluation of 4 popular LLMs in 6 diverse biomedical tasks across 26 datasets has been conducted}. To the best of our knowledge, this is the first work that conducts an extensive evaluation and comparison of various LLMs in the biomedical domain.  Interestingly, we find based on our evaluation that in biomedical datasets that have smaller training sets, {zero-shot LLMs even outperform the current state-of-the-art models when they were fine-tuned only on the training set of these datasets}. This suggests that pre-training on large text corpora makes LLMs quite specialized even in the biomedical domain. We also find that not a single LLM can outperform other LLMs in all tasks, with the performance of different LLMs may vary depending on the task. While their performance is still quite poor in comparison to the biomedical models that were fine-tuned on large training sets, our findings demonstrate that LLMs have the potential to be a valuable tool for various biomedical tasks that lack large annotated data. 
\end{abstract}

\section{Introduction}

The rapid growth of language models \cite{rogers2021primer} in the field of Natural Language Processing (NLP) in recent years has led to significant advancements in various domains, including the biomedical domain \cite{kalyan2022ammu}.  Although specialized models like BioBERT (\textbf{B}idirectional \textbf{E}ncoder \textbf{R}epresentations from \textbf{T}ransformers for \textbf{Bio}medical Text Mining) \cite{lee2020biobert}, BioBART (\textbf{B}idirectional and \textbf{A}uto-\textbf{R}egressive
\textbf{T}ransformers for the \textbf{Bio}medical Domain) \cite{yuan2022biobart}, and BioGPT (\textbf{G}enerative \textbf{P}re-trained \textbf{T}ransformer for \textbf{Bio}medical Text Generation and Mining) \cite{luo2022biogpt} have shown promising results in the biomedical domain, they require fine-tuning\footnote{Fine-tuning means providing good amount (e.g., thousands of samples) of training examples to re-train a pre-trained language model on a specific task.} using domain-specific datasets. This fine-tuning process can be time-consuming due to the requirement of task-specific large annotated datasets. In contrast, zero-shot\footnote{Zero-shot learning means asking a trained model to complete a task without providing any explicit examples of that particular task.} learning \cite{wang2019zeroshotsurvey} enables models to perform tasks without the need for fine-tuning on task-specific datasets. 

\textcolor{black}  {\textbf{L}arge \textbf{L}anguage \textbf{M}odel\textbf{s} (LLMs)} \cite{zhao2023survey} are a class of natural language processing models that have been trained on vast amounts of textual data, making it possible to understand and generate human-like language. In recent years, LLMs such as ChatGPT\footnote{\url{https://chat.openai.com/}} have demonstrated impressive performance on a range of language tasks, including text classification, question answering, and text summarization. One area where LLMs are not yet deeply investigated is the biomedical text processing and information retrieval domain. While there are vast amount of textual data available in the field of biomedicine, there still remains a scarcity of annotated datasets in this domain. Thus, it is difficult to build suitable models for biomedical tasks that lack large annotated datasets. In this regard, due to the strong zero-shot capabilities of LLMs across various tasks, LLM-powered automated tools can be useful for researchers and practitioners in the biomedical domain to find relevant information and extract insights from this vast corpus of unannotated data. However, despite being evaluated on various traditional NLP tasks, there is a lack of comprehensive studies that evaluate LLMs in the biomedical domain. To this end, this paper aims to evaluate LLMs across benchmark biomedical tasks. 

However, the evaluation of LLMs in the biomedical domain would require a proper understanding of the complex linguistic characteristics of biomedical texts. In addition, LLMs are sensitive to prompts \cite{liu2023prompt,jahan-etal-2023-evaluation}. Thus, for biomedical tasks, the effective construction of prompts is important to best utilize these LLMs in biomedical applications. Under these circumstances, domain-specific knowledge in the biomedical domain could play a pivotal role in improving the performance of LLMs in biomedical tasks. In this regard, we study how to effectively build prompts for LLMs to simulate common tasks in biomedical research, such as document classification, named entity recognition, relation extraction, text summarization, question answering, etc.

{Since technologies in medicine and healthcare are critical, it is important to ensure rigorous evaluation before using LLMs in these domains. Thus, this paper will contribute to the understanding of the capabilities and limitations of LLMs in biomedical text processing and information retrieval.} Moreover, with a comprehensive evaluation of various powerful LLMs, this paper would lead to the development of new tools and techniques for researchers in this field, which could pave the way to build new applications in healthcare and biomedicine via leveraging LLMs. The major contributions from this study
are summarized below:

\begin{itemize}
    \item A comprehensive evaluation of various LLMs in the biomedical domain, providing insights into their capabilities and limitations for various tasks. \textcolor{black} {More specifically, this study investigates the zero-shot capabilities of LLMs in the Biomedical domain to address the lack of large annotated datasets in this domain}. 
    \item Construction of task-specific prompts by understanding the complex linguistic structure of biomedical texts. Our findings based on the extensive performance analysis of LLMs across various biomedical tasks will help researchers and practitioners when building LLM-based applications for the biomedical domain.
    \item \textcolor{black}  {To pave the way for future research on LLMs in the biomedical domain, we will release the code used for pre-processing and parsing of LLM-generated responses, alongside the data (the prompts constructed for LLMs and the LLM-generated responses) here: \url{https://github.com/tahmedge/llm-eval-biomed}.}
\end{itemize}




\section{Related Work}
There are a large number of studies on various biomedical tasks, such as biomedical image analysis \cite{liu2023recentimage,rahman2021exploring,morid2021scoping}, 
biomedical text processing \cite{cohen2005survey,wang2021pre}, genomic sequence analysis \cite{peng2018overview,ji2021dnabert}, disease diagnosis \cite{ali2021heart}, 
drug discovery \cite{shaker2021silico,martinelli2022generative,pandiyan2022comprehensive}, cancer research \cite{nguyen2019wnt}, vaccine development \cite{soleymani2022overview}, etc. 
Biomedical text processing is closely related to these tasks as it serves as a critical component and enabler by providing automated methods for extracting information from the vast amount of textual data in the biomedical domain. In this section, we mainly review the existing state-of-the-art approaches for processing large amounts of biomedical textual data, that are the most related to our research.
In the following, we first briefly review various language models used in recent years in the biomedical domain, \textcolor{black} {followed by a brief review of the LLMs that have been studied in this paper}. 

\subsection{Language Models for the Biomedical Domain}

In recent years, the effective utilization of transformer-based \cite{vaswani2017attention} NLP models like BERT \cite{devlin2018bert} and GPT \cite{gpt2} have led to significant progress in the biomedical domain \cite{lee2020biobert,clinicalbert,beltagy2019scibert,blurb,blue,raj2021bioelectra}. BERT leverages the encoder of the transformer architecture, while GPT leverages the decoder of the transformer. In addition to these models, sequence-to-sequence models like BART \cite{bart} that leverage both the encoder and the decoder of the transformer have also emerged as a powerful approach in various text generation tasks in the biomedical domain \cite{yuan2022biobart}. It has been observed that domain-specific pre-training of these models on the biomedical text corpora followed by fine-tuning on task-specific biomedical datasets have helped these models to achieve state-of-the-art performance in a variety of Biomedical NLP (BioNLP) tasks \cite{gu2021domain}. This led to the development of various language models for the biomedical domain, such as BioBERT \cite{lee2020biobert}, ClinicalBERT \cite{clinicalbert}, BioBART \cite{yuan2022biobart}, BioElectra \cite{raj2021bioelectra}, BioGPT \cite{luo2022biogpt}, etc. However, one major limitation of using such fine-tuned models is that they require task-specific large annotated datasets, which is significantly less available in the BioNLP domain in comparison to the general NLP domain. In this regard, having a strong zero-shot model could potentially alleviate the need for large annotated datasets, as it could enable the model to perform well on tasks that it was not exclusively trained on. 

\subsection{Large Language Models}

In recent years, large autoregressive decoder-based language models like GPT-3 \cite{gpt3} have demonstrated impressive few-shot learning capability. With the success of GPT-3 in few-shot scenarios, a new variant of GPT-3 called the InstructGPT model \cite{ouyang2022training} has been proposed that leverages the reinforcement learning \cite{kaelbling1996reinforcement} from human feedback (RLHF) mechanism. The resulting InstructGPT models (in other words, GPT-3.5) are much better at following instructions than the original GPT-3 model, resulting in an impressive zero-shot performance across various tasks. ChatGPT\footnote{\url{https://openai.com/blog/chatgpt}} is the latest addition in the GPT-3.5 series models that additionally uses dialog-based instructional data during its training phase. Recently, more decoder-based \textcolor{black} {LLMs} such as PaLM\footnote{\url{https://ai.google/discover/palm2/}} \cite{chowdhery2022palm,anil2023palm2,singhal2022largemedpalm}, Claude\footnote{\url{https://www.claudeai.ai/}}, LLaMA\footnote{\url{https://ai.meta.com/blog/large-language-model-llama-meta-ai/}} \cite{touvron2023llama,touvron2023llama2} etc. have been proposed that also achieve impressive performance in a wide range of tasks. All these LLMs including ChatGPT are first pre-trained on a large amount of textual data to predict the next token and then fine-tuned using a process called reinforcement learning from human feedback (RLHF) that leveraged both supervised learning and reinforcement learning techniques. The goal of RLHF was to improve the model's performance and ensure that it provided high-quality responses to user queries. The supervised learning phase of the RLHF process involved training the model on conversations in which human trainers played both sides: the user and the AI assistant. These conversations were collected from a variety of sources, including chat logs from customer service interactions, social media messages, and chatbots. The supervised learning phase aimed to train the model to produce high-quality responses that were contextually relevant to the user's query.
Meanwhile, the reinforcement learning phase of the RLHF process aimed to further improve the model's performance by using human trainers to provide feedback on its responses. In this phase, human trainers ranked the responses that the model had created in a previous conversation. These rankings were used to create ``reward models'' that were used to fine-tune the model further by using several iterations of Proximal Policy Optimization (PPO) \cite{kaelbling1996reinforcement}.

While these models have demonstrated strong performance in various NLP tasks \cite{qin2023chatgpt,bang2023multitaskchatgpt,yang2023exploringchatgpt}, they have not been investigated in the biomedical domain yet. To this end, this paper aims to evaluate these powerful LLMs in the biomedical domain. 


\section{Biomedical Tasks Description}

{The biomedical text processing task refers to the use of computational techniques to analyze and extract information from textual data in the field of biomedicine. It can be defined as follows:}

\begin{equation}
   {T: X \rightarrow Y}
\end{equation}

{Here, $X$ represents the input text for the given task $T$, and $Y$ represents the output generated.} {In the following, the description of the benchmark biomedical text processing tasks that have been studied in this paper along with some examples are demonstrated}. 


 \paragraph{\textbf{\textcolor{black} {(i) Biomedical Named Entity Recognition:}}} {Named Entity Recognition (NER) is the task of identifying named entities like person, location, organization, drug, disease, etc. in a given text \cite{yadav-bethard-2018-survey-ner}}. In the case of biomedical NER, this task aims to extract the biomedical named entities, such as genes, proteins, diseases, chemicals, etc., from the literature to improve biomedical research.  
  
\textit{\textbf{Example:} The patient has been diagnosed with a rare form of cancer and is undergoing chemotherapy treatment with the drug Taxol.}

\textit{\textbf{Expected NER classifications: }
\begin{itemize}
    \item NER (Disease): ``rare form of cancer''.
\item NER (Treatment): ``chemotherapy''.
\item NER (Drug): ``Taxol''.
\end{itemize}}

 \paragraph{\textbf{\textcolor{black} {(ii) Biomedical Relation Extraction:}}} {The relation extraction task aims to extract relations between
named entities in a given text \cite{zhong2021frustratingly-re}. In the biomedical relation extraction task, the aim is to analyze textual data by identifying which gene/variants are responsible for which diseases, which treatment/drug is effective for which disease, as well as identifying drug-drug interactions, etc.} 
  
  \textit{\textbf{Example:} The patient has been diagnosed with a rare form of cancer and is undergoing chemotherapy treatment with the drug Taxol.}

\textit{\textbf{Expected Relation Extractions: }
\begin{itemize}
    \item {Relation (Treatment of a Disease):} ``chemotherapy'' is a treatment for ``rare form of cancer''. 
    \item {Relation (Drug used in Treatment):} ``Taxol'' is a drug used in ``chemotherapy''.
\end{itemize}} 

 \paragraph{\textbf{\textcolor{black} {(iii) Biomedical Entity Linking:}}} {The entity linking task focuses on linking named entities in a text to their corresponding entries in a knowledge base \cite{laskar2022auto,laskar2022blink}. In the case of the biomedical entity linking task, it involves recognizing and linking biomedical named entities in unstructured text to their correct definitions, e.g., to the corresponding entries in structured knowledge bases or ontologies.}

\textit{\textbf{Example:} The patient has been diagnosed with a rare form of cancer and is undergoing chemotherapy treatment with the drug Taxol.}

\textit{\textbf{Expected Entity Linking: }
A biomedical entity linking system may link the drug Taxol to the following link: \textit{\url{https://chemocare.com/druginfo/taxol}.}} 


 \paragraph{\textbf{\textcolor{black} {(iv) Biomedical Text Classification:}}} For a given text, the goal of this task is to classify the text into a specific category. One example to classify a given sentence in one of the 10 hallmarks of cancer taxonomy has been demonstrated below: 

\textit{\textbf{Example:}} ``Heterogeneity in DNA damage within the cell population was observed as a function of radiation dose.''

 \textit{\textbf{Expected Result:}} Genomic Instability and Mutation.

 \paragraph{\textbf{\textcolor{black} {(v) Biomedical Question Answering:}}} 
The biomedical question-answering task involves retrieving the relevant answer for the given question related to the biomedical literature, such as scientific articles, medical records, and clinical trials. This task is of great importance as it can help healthcare professionals, researchers, and patients access relevant information quickly and efficiently, which can have a significant impact on patient care, drug development, and medical research.

\textit{\textbf{Example:} What is recommended for thalassemia patients
?}

\begin{itemize}
    \item \textit{Candidate Answer 1: Chemotherapy may be used to: Cure the cancer, shrink the cancer, and prevent the cancer from spreading.}

    \item \textit{Candidate Answer 2: Regular blood transfusions can help provide the body with normal red blood cells containing normal hemoglobin.}

\end{itemize}

\textit{\textbf{Expected Answer:}} The candidate answer 2 should be retrieved as a relevant answer \cite{abacha2019overview,he-etal-2020-infusing-mediqa2019}.

 \paragraph{\textbf{\textcolor{black} {(vi) Biomedical Text Summarization:}}}{The main purpose of the text summarization task is to generate a short concise summary of the given document \cite{el2021automatic}. The generation of short summaries of biomedical texts would help reduce the time spent reviewing lengthy electronic health records / patient queries in healthcare forums / doctor-patient conversations, resulting in improving the efficiency of the healthcare system.} 
  
\textit{\textbf{Example:} Patient is a 62-year-old female with a medical history of hyperlipidemia, osteoarthritis, and previous cerebrovascular accident. She presented with sudden onset of dizziness and palpitations that began a day ago. An electrocardiogram was immediately conducted, which indicated the presence of atrial fibrillation. She was promptly hospitalized for monitoring and commenced on anticoagulation therapy with warfarin and rate-controlling medications like beta-blockers.}

\textit{\textbf{Expected Summary:} A 62-year-old female with a history of hyperlipidemia, osteoarthritis, and a previous cerebrovascular accident experienced sudden dizziness and palpitations. An ECG confirmed atrial fibrillation, leading to her hospitalization and treatment with warfarin and beta-blockers.}


\section{Methodology}

\begin{figure*}[t!]
	\centering
		\includegraphics[scale=.37]{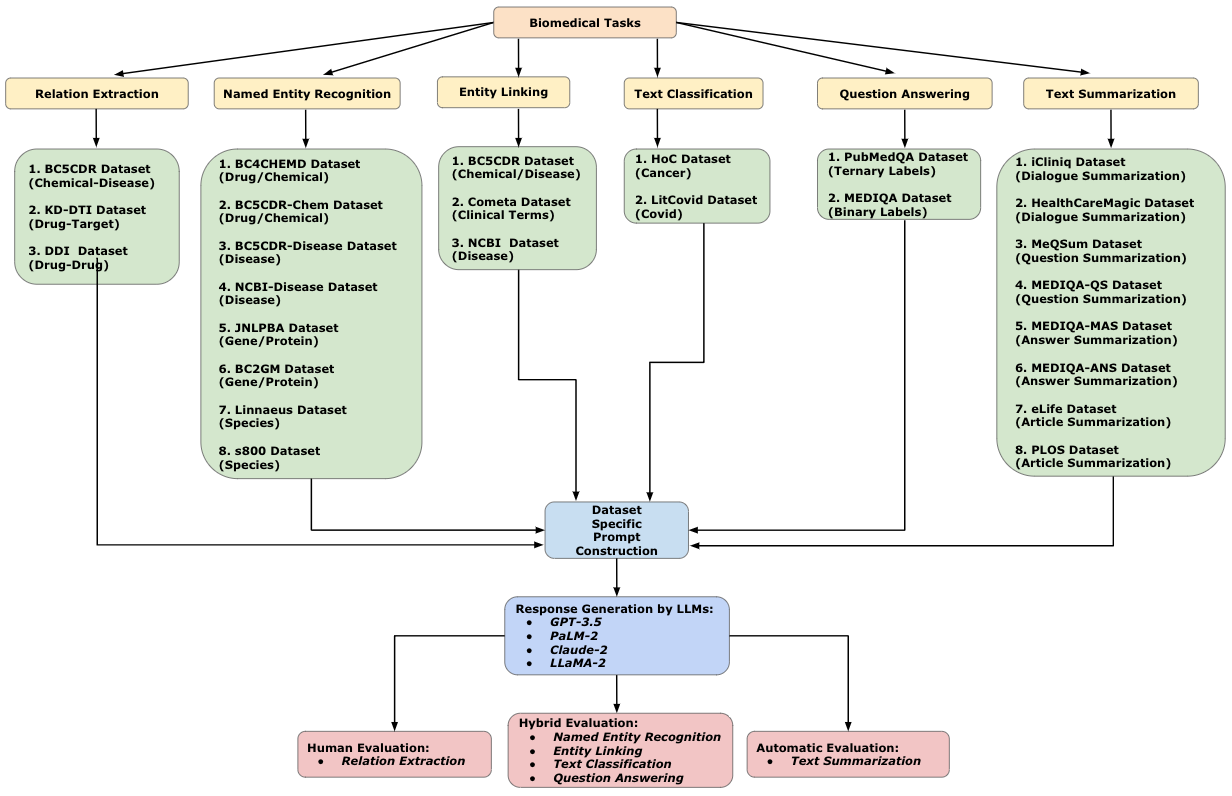}
	\caption{\textcolor{black} {An overview of our methodology to evaluate 6 biomedical tasks across 26 datasets in this paper. At first, we construct the prompt for each dataset. Then, we generate the response for each dataset using respective LLMs. Finally, depending on the task, we apply various evaluation techniques.}}
	\label{fig:overview}
\end{figure*}

\textcolor{black} {In this section, we first present our methodology on how we design the prompts for different tasks, followed by describing the LLMs that have been studied in this paper. Afterward, the evaluation pipeline has been demonstrated. An overview of our methodology is also shown in Figure \ref{fig:overview}.} 

\subsection{Prompt Design}\label{prompt_design}

For a given test sample $X$, we first prepare a task instruction $T$. Then, we concatenate the test sample $X$ with the task instruction $T$ to construct the prompt $P$. Afterward, the prompt $P$ is given as input to generate the response $R$. \textcolor{black} {Below, the prompt $P$ that has been constructed for each task depending on the respective dataset has been demonstrated}.

\paragraph{\textbf{(i) NER:}} For NER, prompts are designed to identify the biomedical named entities in a given text in the BIO format. In our prompts, the description of the BIO format is also added along with the task instructions. For NER, we use the BC2GM \cite{bc2gm} and JNLPBA \cite{jnlpba} datasets for gene/protein entity recognition, BC4CHEMD \cite{bc4chemd} and BC5CDR-CHEM \cite{bc5cdr} for drug/chemical entity recognition, BC5CDR-Disease \cite{bc5cdr} and NCBI-Disease \cite{ncbi} for disease type entity recognition, LINNAEUS \cite{linnaeus} and s800 \cite{s800} for species type entity recognition. The prompts for this task are shown in Table \ref{tab:input_prompt_ner}.

\paragraph{\textbf{(ii) Relation Extraction:}} 
To identify the possible relation between entities mentioned in a given text, the prompts are designed depending on the dataset. For this purpose, we construct prompts for chemical-disease-relation in the BC5CDR dataset \cite{bc5cdr}, drug-target-interaction in the KD-DTI dataset \cite{kddti}, and drug-drug-interaction in the DDI dataset \cite{ddi}. The prompts used for these datasets are demonstrated in Table \ref{tab:input_prompt_relation_extraction}.

\paragraph{\textbf{(iii) Entity Linking:}} 
To identify whether LLMs can link named entities to their correct definitions based on their pre-training knowledge, we follow the work of Yuan et al. \cite{yuan2022generativeentitylinking} for the generative entity linking task by asking LLMs to identify the correct concept names for the named entities. \textcolor{black} {For evaluation, the BC5CDR \cite{bc5cdr} dataset for the entity linking of disease/chemical type named entities, the NCBI \cite{ncbi} dataset to link diseases, and the COMETA \cite{cometa} dataset to link clinical terms have been used}. The sample prompts for this task are shown in Table \ref{tab:input_prompt_el}.

\begin{table*}[t!]
\centering
\tiny

\caption{{Sample Prompts in Different Named Entity Recognition (NER) Datasets.}}
\begin{tabular}{p{1.5cm}p{2.5cm}p{2cm}p{8cm}}
\toprule
\textbf{Dataset} & \textbf{Type} & \textbf{Data Split \newline (Train / Valid / Test)} & \textbf{Prompt} \\
\midrule

BC2GM \newline BC4CHEMD \newline BC5CDR-CHEM \newline BC5CDR-Disease \newline JNLPBA \newline LINNAEUS 
 \newline NCBI-Disease \newline s800 & 
 NER (GENE/PROTEIN) \newline 
 NER (DRUG/CHEMICAL) \newline 
NER (DRUG/CHEMICAL) \newline 
NER (DISEASE) \newline 
NER (GENE/PROTEIN) \newline 
NER (SPECIES) \newline  
NER (DISEASE) \newline  
NER (SPECIES)
 & 12574 / 2519 / 5038 \newline
 30682 / 30639 / 26364 \newline
 4560 / 4581 / 4797 \newline 
 4560 / 4581 / 4797 \newline 
 14690 / 3856 / 3856 \newline
 11935 / 4078 / 7142 \newline 
 5424 / 923 / 940 \newline 
 5733 / 830 / 1630 
 &
Below, we provide a biomedical text:

[TEXT]

You need to identify the [ENTITY] type named entities in the above text.
To identify the named entities, please tag each token of the given text in the 'BIO' format as either: 'B' or 'I' or 'O'
The BIO format stands for Beginning, Inside, Outside. It provides a way to label individual tokens in a given text to indicate whether they are part of a named entity.
In the BIO format, each token in a text is labeled with a tag that represents its role in a named entity. For our case, there are three possible tags:
B: it indicates that the token is the beginning of the [ENTITY] type named entity (i.e., the first token of a [ENTITY] type named entity).
I: it indicates the token is inside a [ENTITY] type named entity (i.e., any token other than the first token of a [ENTITY] type named entity).
O: it indicates that the token is outside any named entity. In other words, it is not part of any named entity.
Below, each token of the biomedical text is provided (separated by new line). Now please assign the correct tag to each token. Return your result for each token in a newline in the following format -> token: assigned\_tag: \newline [LIST OF LINE SEPARATED TOKENS]
\\ 
\bottomrule
\end{tabular}

\label{tab:input_prompt_ner}
\end{table*}

\begin{table*}[t!]
\centering
\tiny

\caption{\small{Sample Prompts in Different Relation Extraction Datasets.}}
\begin{tabular}{p{1cm}p{2cm}p{2cm}p{9cm}}

\toprule
\textbf{Dataset} & \textbf{Type} & \textbf{Data Split \newline (Train / Valid / Test)} & \textbf{Prompt} \\
\midrule

BC5CDR & Chemical-Disease \newline Relation Extraction & 500 / 500 / 500 & Identify each pair of drugs and the drug-induced side-effects (e.g., diseases)  in the following passage: \newline [PASSAGE]
\\ 
\midrule
KD-DTI & Drug-Target \newline Relation Extraction & 12K / 1K / 1.3K & Identify the drug-target interactions in the following passage (along with the interaction type among the following: 'inhibitor', 'agonist', 'modulator', 'activator', 'blocker', 'inducer', 'antagonist', 'cleavage', 'disruption', 'intercalation', 'inactivator', 'bind', 'binder', 'partial agonist', 'cofactor', 'substrate', 'ligand', 'chelator', 'downregulator', 'other', 'antibody', 'other/unknown'): [PASSAGE]
\\ 
\midrule
DDI & Drug-Drug \newline Relation Extraction & 664 / 50 / 191 & Identify the pairs of drug-drug interactions in the passage given below based on one of the following interaction types: \newline
(i) mechanism: this type is used to identify drug-drug interactions that are described by their pharmacokinetic mechanism. \newline
(ii) effect: this type is used to identify drug-drug interactions describing an effect. \newline  
(iii) advice: this type is used when a recommendation or advice regarding a drug-drug interaction is given. \newline 
(iv) int: this type is used when a drug-drug interaction appears in the text without providing any additional information. \newline [PASSAGE] \\
\bottomrule
\end{tabular}
\label{tab:input_prompt_relation_extraction}
\end{table*}

\begin{table*}[t!]
\centering
\tiny

\caption{{Sample Prompts in Different Entity Linking Datasets.}}
\begin{tabular}{p{2cm}p{4cm}p{3cm}p{5cm}}
\toprule
\textbf{Dataset} & \textbf{Type} & \textbf{Data Split \newline (Train / Valid / Test)} & \textbf{Prompt} \\
\midrule
BC5CDR \newline COMETA \newline NCBI & 
Entity Linking (DISEASE/CHEMICAL) \newline 
Entity Linking (CLINICAL TERMS) \newline 
Entity Linking (DISEASE) & 
9285 / 9515 / 9654 
\newline 13489 / 2176 / 4350 
\newline 5784 / 787 / 960 &

[TEXT\_S <START> ENTITY <END> TEXT\_E]

In the biomedical text given above, what does the entity between the START and the END token refer to?\\

\bottomrule
\end{tabular}

\label{tab:input_prompt_el}
\end{table*}

\begin{table*}[t!]
\centering
\tiny
\caption{{Sample Prompts in Different Text Classification Datasets.}}
\begin{tabular}{p{1.5cm}p{2.5cm}p{2.5cm}p{7.5cm}}
\toprule
\textbf{Dataset} & \textbf{Type} & \textbf{Data Split \newline (Train / Valid / Test)} & \textbf{Prompt} \\
\midrule
HoC & Text Classification & 9972 / 4947 / 4947 &

The 10 hallmarks of cancer taxonomy with their definitions are given below: \newline
(i) Sustaining proliferative signaling: Cancer cells can initiate and maintain continuous cell division by producing their own growth factors or by altering the sensitivity of receptors to growth factors. \newline
(ii) Evading growth suppressors: Cancer cells can bypass the normal cellular mechanisms that limit cell division and growth, such as the inactivation of tumor suppressor genes and/or insensitivity to antigrowth signals. \newline
(iii) Resisting cell death: Cancer cells develop resistance to apoptosis, the programmed cell death process, which allows them to survive and continue dividing. \newline
(iv) Enabling replicative immortality: Cancer cells can extend their ability to divide indefinitely by maintaining the length of telomeres, the protective end caps on chromosomes. \newline
(v) Inducing angiogenesis: Cancer cells stimulate the growth of new blood vessels, providing the necessary nutrients and oxygen to support their rapid growth. \newline
(vi) Activating invasion and metastasis: Cancer cells can invade surrounding tissues and migrate to distant sites in the body, forming secondary tumors called metastases. \newline
(vii) Deregulating cellular energetic metabolism: Cancer cells rewire their metabolism to support rapid cell division and growth, often relying more on glycolysis even in the presence of oxygen (a phenomenon known as the Warburg effect). \newline
(viii) Avoiding immune destruction: Cancer cells can avoid detection and elimination by the immune system through various mechanisms, such as downregulating cell surface markers or producing immunosuppressive signals. \newline
(ix) Tumor promoting inflammation: Chronic inflammation can promote the development and progression of cancer by supplying growth factors, survival signals, and other molecules that facilitate cancer cell proliferation and survival. \newline
(x) Genome instability and mutation: Cancer cells exhibit increased genomic instability, leading to a higher mutation rate, which in turn drives the initiation and progression of cancer. \newline

Classify the sentence given below in one of the above 10 hallmarks of cancer taxonomy (if relevant). If cannot be classified, answer as ``empty": \newline
[SENTENCE]
\\ 
\midrule
LitCovid & Text Classification & 16126 / 2305 / 4607 &
Choose the most appropriate topic(s) for the biomedical article on covid-19 given below from the following options: (i) Prevention, (ii) Treatment, (iii) Diagnosis, (iv) Mechanism, (v) Case Report, (vi) Transmission, (vii) Forecasting, and (viii) General.

[ARTICLE]
\\ 
\bottomrule
\end{tabular}
\label{tab:input_prompt_text_classification}
\end{table*}

\paragraph{\textbf{(iv) Text Classification:}} 
The goal of this task is to classify the type of the given text. 
In this paper, we use two datasets: (i) the HoC (the Hallmarks of Cancer corpus) dataset \cite{hoc}, and (ii) the LitCovid dataset \cite{chen2021litcovid}. The HoC dataset consists of 1580 PubMed abstracts where the goal is to annotate each sentence in the given abstract in one of the 10 currently known hallmarks of cancer. Whereas in the LitCovid dataset, each article is required to be classified in one (or more) of the following 8 categories: Prevention, Treatment, Diagnosis, Mechanism, Case Report, Transmission, Forecasting, and General. 
Our prompts for these text classification datasets are shown in Table \ref{tab:input_prompt_text_classification}.

\paragraph{\textbf{(v) Question Answering:}} 
For the question-answering task, we also evaluate the performance of LLMs on multiple datasets: (i) the PubMedQA dataset \cite{jin2019pubmedqa}, and (ii) the MEDIQA-2019 dataset \cite{abacha2019overview}. In the PubmedQA dataset, the question, the reference context, and the answer are given as input to the LLMs to determine whether the answer to the given question can be inferred from the provided reference context with LLMs being prompted to reply either as \textit{yes}, \textit{no}, or \textit{maybe}, as required by the task. In the MEDIQA-2019 dataset, the LLMs are asked to determine whether the retrieved answer for the given question is relevant or not \cite{laskar-LREC}. The prompts for this task are shown in Table \ref{tab:input_prompt_qa}. 

\begin{table*}[t!]
\centering
\tiny
\caption{{Sample Prompts in Different Question Answering Datasets.}}
\begin{tabular}{p{1.5cm}p{2cm}p{2.5cm}p{8cm}}
\toprule
\textbf{Dataset} & \textbf{Type} & \textbf{Data Split \newline (Train / Valid / Test)} & \textbf{Prompt} \\
\midrule
PubMedQA & Question \newline Answering & 450 / 50 / 500 & For the question, the reference context, and the answer given below, is it possible to infer the answer for that question from the reference context? Only reply as either Yes or No or Maybe.

Question: [QUESTION] 

Reference context:  [REFERENCE CONTEXT]

Answer:  [ANSWER]
\\

\midrule
MEDIQA-2019 & Question \newline Answering & {1701 / 234 / 1107} & 

A retrieved answer for the following question is given below. Identify whether the retrieved answer is relevant to the question or not. Answer as 1 if relevant, otherwise answer as 0.

Question: [QUESTION]

Retrieved Answer: [TEXT]
\\ 
\bottomrule
\end{tabular}
\label{tab:input_prompt_qa}
\end{table*}

\paragraph{\textbf{(vi) Text Summarization:}} The biomedical text summarization task requires the generation of a concise summary of the given biomedical text. To this end, the LLMs are evaluated across a wide range of diverse biomedical summarization tasks, such as healthcare question summarization (\textit{MeQSum} \cite{abacha2019summarization} and \textit{MEDIQA-QS} \cite{abacha2021overview} datasets), medical answer summarization (\textit{MEDIQA-ANS} \cite{savery2020questionmediqaans} and \textit{MEDIQA-MAS} \cite{abacha2021overview} datasets), and doctor-patient dialogue summarization (\textit{iCliniq} and \textit{HealthCareMagic} datasets \cite{zeng2020meddialog,mrini2021gradually}) to generate short queries for healthcare forums describing patient's medical conditions. In addition, we use various datasets for biomedical literature summarization \cite{luo-etal-2022-readability,goldsack-etal-2022-making}, such as the Biomedical Text Lay Summarization shared task 2023 (BioLaySumm-2023) datasets \cite{goldsack2023biolaysumm}. For BioLaySumm-2023, since the gold reference summaries of the test sets are not publicly available as of the writing of this paper, \textcolor{black} {the respective validation sets are used for evaluation}. 
The sample prompts in the summarization datasets are shown in Table \ref{tab:input_prompt_summ}.

\begin{table*}[t!]
\centering
\tiny
\caption{Sample Prompts in Different Text Summarization tasks.}
\begin{tabular}{p{3cm}p{3.5cm}p{3cm}p{4.75cm}}
\toprule
\textbf{Dataset} & \textbf{Type} & \textbf{Data Split \newline(Train / Valid / Test)} &  \textbf{Prompt} \\
\midrule
iCliniq & Dialog \newline Summarization & 24851 / 3105 / 3108 & Write a very short and concise one line summary of the following dialogue as an informal question in a healthcare forum:
 \newline [DIALOGUE]
\\ 
\midrule
HealthCare Magic & Dialog \newline  Summarization & 181122 / 22641 / 22642 & Write a very short and concise one line summary of the following dialogue as a question in a healthcare forum:
  \newline [DIALOGUE]
\\ 
\midrule
MeQSum & Question \newline  Summarization & 500 / - / 500 & Rewrite the following question in a short and concise form:   \newline [QUESTION]
\\
\midrule
MEDIQA-QS & Question  \newline Summarization & - / 50 / 100 & Rewrite the following question in a short and concise form:   \newline [QUESTION]
\\

\midrule
MEDIQA-MAS & Answer  \newline Summarization & - / 50 / 80 & For the following question, some relevant answers are given below. Please write down a short concise answer by summarizing the given answers. 

Question: [QUESTION]

Answer 1:  [ANSWER1]

Answer 2:  [ANSWER2]
\\
\midrule
MEDIQA-ANS & Answer \newline  Summarization & - / - / 552 & Write a very short and concise summary of the following article based on the question given below:  \newline [QUESTION]  \newline [ARTICLE]
\\
\midrule
BioLaySumm-2023 (PLOS) & Lay \newline Summarization  & 24773 / 1376 / 142 & Write down a readable summary of the following biomedical article using less technical terminology (e.g., lay summary) such that it can be understandable for non-expert audiences:  \newline [ABSTRACT + ARTICLE]
\\
\midrule
BioLaySumm-2023 (eLife) & Lay \newline Summarization  & 4346 / 241 / 142 & Write down a readable summary of the following biomedical article using less technical terminology (e.g., lay summary) such that it can be understandable for non-expert audiences:  \newline [ABSTRACT + ARTICLE]
\\
\midrule
BioLaySumm-2023 (PLOS) & Readability-controlled  \newline Summarization  (Lay Summary) & 24773 / 1376 / 142 & Write down a readable summary of the following biomedical article using less technical terminology (e.g., lay summary) such that it can be understandable for non-expert audiences:  \newline [ARTICLE]
\\
\midrule
BioLaySumm-2023 (PLOS) & Readability-controlled \newline Summarization  (Abstract) & 24773 / 1376 / 142 & Write down the abstract of the following biomedical article:  \newline [ARTICLE]

\\
\bottomrule

\end{tabular}

\label{tab:input_prompt_summ}
\end{table*}

\subsection{Models}

In the following, we describe the 4 popular LLMs that we evaluate in benchmark biomedical datasets and tasks in this paper. 

\paragraph{\textbf{(i) GPT-3.5:}} GPT-3.5 is an auto-regressive language model based on the transformer \cite{vaswani2017attention} architecture that was pre-trained on a vast amount of textual data via supervised learning alongside reinforcement learning with human feedback. The backbone model behind the first version of ChatGPT was also GPT-3.5, and it is currently one of the base models, behind OpenAI's ChatGPT, alongside GPT-4. The initial training data for GPT-3.5 was obtained from a large corpus of text data that was crawled from the internet. This corpus included a wide range of publicly available text, including articles, books, and websites. Additionally, OpenAI collected data from GPT-3 users to train and fine-tune the model further \cite{qin2023chatgpt,openai2023gpt4}. In this work, we used the OpenAI API for the \textit{gpt-3.5-turbo-0613\footnote{\url{https://platform.openai.com/docs/models/gpt-3-5}}} model for GPT-3.5. 

\paragraph{\textbf{(ii) PaLM-2:}} PaLM-2 \cite{anil2023palm2} is also a transformer-based language model that exhibits enhanced multilingual and reasoning capabilities, along with improved computing efficiency. It is the base model behind Google's BARD\footnote{\url{https://bard.google.com/}}, which is a competitor to OpenAI's ChatGPT. The computational efficiency in PaLM-2 is achieved by scaling the model size and the training dataset size in proportion to each other. This new technique makes PaLM-2 smaller than its predecessor, PaLM-1, while achieving better performance, including faster inference, fewer parameters to serve, and a lower serving cost. It is trained using a mixture of objectives, allowing it to learn various aspects of language and reasoning across a diverse set of tasks and capabilities, making it a powerful tool for various applications. In this work, we used the \textit{text-bison@001} model in Google's Vertex AI\footnote{\url{https://cloud.google.com/vertex-ai/docs/generative-ai/model-reference/text}} API for PaLM-2. 

\paragraph{\textbf{(iii) Claude-2:}} Claude-2 is also a general-purpose LLM based on the transformer architecture. It was developed by Anthropic\footnote{\url{https://www.anthropic.com/index/claude-2}} and is a successor of Claude-1. Similar to other large models, it is trained via unsupervised pre-training, supervised fine-tuning, and reinforcement learning with human feedback. Internal red-teaming evaluation by Anthropic shows that Claude is more harmless and less likely to produce offensive or dangerous output. Experimental evaluation of Claude-1 and Claude-2 demonstrates that Claude-2 achieves much better performance than Claude-1 across various tasks. Thus, we also utilize Claude-2 in this work via leveraging Anthropic's \textit{claude-2} API. 

\paragraph{\textbf{(iv) LLaMA-2:}} LLaMA-2 \cite{touvron2023llama2} is a recently proposed LLM by Meta\footnote{\url{https://ai.meta.com/llama/}}. One major advantage of LLaMA-2 over the previously mentioned LLMs is that it is also open-sourced. While another open-sourced version of LLaMA: the LLaMA-1 \cite{touvron2023llama} model was released prior to the release of LLaMA-2, the LLaMA-1 model was only allowed for non-commercial usage. On the contrary, the recently proposed LLaMA-2 not only allows commercial usage, but also outperforms its earlier open-sourced version LLaMA-1 across a wide range of tasks.
This makes LLaMA-2 a breakthrough model in both academia and industry. Similar to other LLMs, LLaMA-2 is also trained via unsupervised pre-training, supervised fine-tuning, and reinforcement learning with human feedback. Note that the LLaMA-2 model has been released in various sizes: 7B, 13B, and 70B. While the 70B model has achieved the best performance across various benchmarks, it requires very high computational resources. On the other hand, although the 7B model requires less computational resources, it achieves poorer performance in comparison to the 13B and 70B models. Considering the performance and cost trade-off, we used the LLaMA-2-13B\footnote{We used the following version of LLaMA-2-13B: \url{https://huggingface.co/meta-llama/Llama-2-13b-chat-hf}, which achieves improved factual correctness than its based version. As we are benchmarking LLMs in the biomedical domain, \textcolor{black} {selecting a more faithful model is prioritized}.} model in this work.

\subsection{Evaluation Pipeline} Since LLMs usually generate human-like responses that may sometimes contain unnecessary information while not in a specific format, some tasks are very challenging to evaluate without any human intervention. For instance, in tasks like Relation Extraction, there can be multiple answers. Thus, it would be very difficult to automatically evaluate the performance of LLMs by comparing their response with the gold labels using just an evaluation script. Thus, in this paper, to ensure high-quality evaluation,  we follow the work of Laskar et al. \cite{laskar-etal-2023-systematic}, where they design different settings for the evaluation of LLMs for different tasks:
\begin{enumerate}[label=\roman*.]
    \item \textbf{Automatic Evaluation:} Where they evaluate some tasks, such as text summarization via leveraging automatic evaluation scripts.
    \item \textbf{Human Evaluation:} Where they evaluate some discriminative tasks solely by humans, which cannot be evaluated directly based on automatic evaluation scripts.
    \item \textbf{Hybrid (Human + Automatic) Evaluation:} Where they evaluate some tasks via leveraging both human intervention alongside evaluation scripts. More specifically, this is done by first applying evaluation scripts on the dataset to parse the results from the LLM-generated response, followed by utilizing human intervention if solely depending on the evaluation script cannot parse the results in the expected format. 
\end{enumerate}

\textbf{\textit{For discriminative tasks}}, where parsing of the results from the generated response is required for evaluation, we follow the work of Laskar et al. \cite{laskar-etal-2023-systematic} and design an evaluation script for the respective dataset to first parse the results and then compare the parsed results with the gold labels. Subsequently, any samples where the script could not parse the result properly were manually reviewed by the human annotators. For NER, Entity Linking, Text Classification, and Question Answering, we evaluate the performance by leveraging this technique (denoted as \textit{hybrid evaluation}). However, for relation extraction, human intervention is necessary since parsing scripts cannot properly identify the relations found in the generative responses. Thus, for relation extraction, all LLM-generated responses were manually evaluated by humans. This technique of solely utilizing humans to evaluate LLM-generated response when parsing is not possible was also used in recent literature \cite{laskar-etal-2023-systematic,jahan-etal-2023-evaluation}. In our human evaluation, at least two annotators compared the LLM-generated response against the gold labels. Any disagreements were resolved based on discussions between the annotators.

\textbf{\textit{For generative tasks}}, such as summarization, where the full response generated by LLMs can be used for evaluation instead of parsing the response, we evaluate using automatic evaluation metrics (e.g., ROUGE or BERTScore).

\section{Experiments}

\subsection{Evaluation Metrics}

{We use different evaluation metrics for different tasks to ensure a fair comparison of different LLMs with prior state-of-the-art results. For this purpose, the standard evaluation metrics that are used in the literature for benchmarking the performance of different models are selected. Thus, for the relation extraction and named entity recognition tasks, Precision, Recall, and F1 metrics are used, while for entity linking, the Recall@1 metric is used. For Summarization, the ROUGE \cite{lin2004rouge} and the BERTScore \cite{zhang2019bertscore} metrics are used. For question answering and text classification, metrics like Accuracy and F1 are used.}

\subsection{Baselines}

{To compare the performance of the zero-shot LLMs, the current state-of-the-art fine-tuned models are used as the baselines. These baseline models are described below.}

    \paragraph{\textbf{\textcolor{black} {(i) BioGPT:}}} The backbone of BioGPT \cite{luo2022biogpt} is GPT-2 \cite{gpt2}, which is a decoder of the transformer \cite{vaswani2017attention}. The BioGPT model was trained over PubMed titles and abstracts via leveraging the standard language modeling task. We use the fine-tuned BioGPT models as the baseline for all datasets in the relation extraction task, HoC dataset in the text classification task, and the PubMedQA\footnote{In PubMedQA, BioGPT was additionally fine-tuned on more than 270K instances.} dataset for the question-answering tasks.

     \paragraph{\textbf{\textcolor{black} {(ii) BioBART:}}} It is a sequence-to-sequence
model based on the BART \cite{bart} architecture where the pre-training process involves reconstructing corrupted input sequences. The main difference between BioBART \cite{yuan2022biobart} and BART is that the former was pre-trained over PubMed abstracts to make it suitable for the biomedical domain tasks. The fine-tuned BioBART model was used as the baseline in all the entity linking datasets and the following biomedical summarization tasks: Dialogue Summarization, Question Summarization, and Answer Summarization.

      \paragraph{\textbf{\textcolor{black} {(iii) BioBERT:}}} It is a domain-specific language representation model \cite{lee2020biobert} based on the BERT \cite{devlin2018bert} architecture that was additionally pre-trained on large-scale biomedical corpora (PubMed and PMC abstracts). The fine-tuned BioBERT model achieved state-of-the-art performance across different biomedical NER datasets and so it was used as the baseline for all NER datasets in this paper. In addition, it was used as the baseline in the LitCovid dataset for text classifcation.

     \paragraph{ \textbf{\textcolor{black} {(iv) ALBERT \textit{with disease knowledge infused}:}}} The ALBERT \cite{albert} model is a variant of the BERT \cite{devlin2018bert} language model which requires lower memory consumption and a new self-supervised loss function. He at al., \cite{he-etal-2020-infusing-mediqa2019} extends its training mechanism by additionally training ALBERT on 14K biomedical texts in a question-answering fashion via infusing disease knowledge which led to the state-of-the-art performance in the MediQA-2019 dataset. The LLMs are compared with this \textit{disease knowledge infused} version of the ALBERT model in this work.

     \paragraph{\textbf{\textcolor{black} {(v) FLAN-T5-XL:} FLAN-T5 \cite{FLAN-t5}}} is an extension of the T5 \cite{t5} model. The T5 model treats each tasks as a sequence to sequence problem. While the architecture of FLAN-T5 is similar to the original T5 model, it leverage instruction fine-tuning instead of traditional fine-tuning. The FLAN-T5-XL that achieves state-of-the-art performance in the Biomedical Lay Summarization task is used as the baseline in the eLife and the PLOS datasets to compare LLMs in biomedical lay summarization.

     \paragraph{\textbf{\textcolor{black} {(vi) PRIMERA:}}} It is a pre-trained model  \cite{xiao-etal-2022-primera} designed to enhance multi-document summarization. It proposes a new pre-training strategy for multi-document summarization by leveraging the longformer-encoder-decoder \cite{beltagy2020longformer} for pre-training. In this work, the fine-tuned PRIMERA model is used as the baseline in the Readability-Controlled Summarization task since it is the current state-of-the-art in this task.

\subsection{Results}

In this section, the results for LLMs in various tasks are presented. At first, we present our results in the Relation Extraction task where we utilize \textit{human evaluation}. Then, we demonstrate our findings in Text Classification, Question Answering, Entity Linking, and NER datasets where \textit{hybrid evaluation }is conducted. Finally, we present our findings in the Summarization datasets where \textit{automatic evaluation} is utilized. 
\begin{table*}[t!]
\small
\centering
\caption{{Performance on Relation Extraction datasets. All SOTA results are taken from the BioGPT \cite {luo2022biogpt} model}.}
\begin{tabular}{c|ccc|ccc|ccc}
\toprule

& \multicolumn{9}{c}{\textbf{Dataset}} \\  

\cmidrule(lr){2-10}

   \textbf{Model}  & \multicolumn{3}{c}{\textbf{BC5CDR}} & \multicolumn{3}{c}{\textbf{KD-DTI}} & \multicolumn{3}{c}{\textbf{DDI}}  \\ 
\cmidrule(lr){2-4}\cmidrule(lr){5-7}\cmidrule(lr){8-10}

& \textbf{Precision} & \textbf{Recall} & \textbf{F1} & \textbf{Precision} & \textbf{Recall} & \textbf{F1} & \textbf{Precision} & \textbf{Recall} & \textbf{F1} \\ 
\midrule

\textbf{GPT-3.5} & {30.62} & {73.85} & {43.29} & 19.19 & {66.02} & 29.74 & \textbf{47.11} & \textbf{45.77} & \textbf{46.43} \\ 

\textbf{PaLM-2} & \textbf{51.61} & {57.30} & \textbf{54.30} & \textbf{40.21} & {36.82} & \textbf{38.44} & {35.47} & {16.48} & {22.50}  \\ 

\textbf{Claude-2} & {44.04} & {67.73} & {53.37} & {17.99} & \textbf{72.73} & 28.84 & {39.27} & {46.60} & {42.62}  \\ 

\textbf{LLaMA-2-13b} & {39.54} & \textbf{81.66} & {53.28} & {15.14} & {60.48} & {24.21} & {22.58} & {25.67} & {24.03}   \\ 

\midrule
\textbf{State-of-the-Art (SOTA)} & 49.52 & 43.25 & 46.17 & {40.00} & 39.72 & {38.42} & 41.70 & 44.75 & 40.76 \\ 
\bottomrule

\end{tabular}

\label{tab:performance_re}
\end{table*}

\begin{table*}[t!]
\small
\setlength{\tabcolsep}{3pt} 
\centering
\caption{{Performance on Text Classification, Question Answering (QA), and Entity Linking datasets. The SOTA results for HoC and PubMedQA are taken from the BioGPT \cite{luo2022biogpt} model, while we take the SOTA results from \citet{gutierrez2020documentclassification} and \citet{he-etal-2020-infusing-mediqa2019} for LitCovid and MediQA-2019, respectively. Note that all SOTA results for Entity Linking are taken from the BioBART \cite{yuan2022biobart} model.}}

\begin{tabular}{c|cc|cc|ccc}
\toprule

& \multicolumn{2}{c}{\textbf{Text Classification Dataset}} & \multicolumn{2}{c}{\textbf{Question Answering Dataset}} & \multicolumn{3}{c}{\textbf{Entity Linking Dataset}}  \\ \cmidrule(lr){2-3}\cmidrule(lr){4-5}\cmidrule(lr){6-8}

\textbf{Model}   & {\textbf{HoC}} & {\textbf{LitCovid}} & {\textbf{PubMedQA}} & {\textbf{MediQA-2019}}& \textbf{BC5CDR} & \textbf{Cometa} & \textbf{NCBI}  \\ 
\cmidrule(lr){2-2}\cmidrule(lr){3-3}\cmidrule(lr){4-4}\cmidrule(lr){5-5} \cmidrule(lr){6-6} \cmidrule(lr){7-7} \cmidrule(lr){8-8}

  & \textbf{F1} & \textbf{F1} & \textbf{Accuracy} & \textbf{Accuracy} & \textbf{Recall@1} & \textbf{Recall@1} & \textbf{Recall@1} \\ 
\midrule
\textbf{GPT-3.5} & {59.26} & 29.63 & 54.40 & \textbf{73.26} & 54.90 & 43.45 & 52.19 \\ 

\textbf{PaLM-2} & \textbf{61.03} & \textbf{37.50} & {59.60} & 52.12 & 52.14 & 48.76 & 38.44  \\ 

\textbf{Claude-2} & 34.93 & 7.60 & 57.20 & 65.13 & \textbf{78.01} & \textbf{53.29} & \textbf{70.21} \\ 

\textbf{LLaMA-2-13b} & 41.82 & 11.34  & \textbf{61.40} & 56.01 & {66.52} & {40.67} & {59.17} \\

\midrule
\textbf{State-of-the-Art (SOTA)} & \textbf{85.12} & \textbf{86.20} & \textbf{78.20} &  \textbf{79.49} &\textbf{93.26}  & \textbf{81.77} & \textbf{89.90} \\ 

\bottomrule

\end{tabular}

\label{tab:performance_tc_qa_el}
\end{table*}

\paragraph{\textbf{(i) Relation Extraction:}} We compare the performance of LLMs with the current state-of-the-art fine-tuned BioGPT \cite{luo2022biogpt} model across 3 datasets for the relation extraction task. {The LLM generated responses in the relation extraction task are computed based on \textbf{\textit{Human Evaluation}}}. From the results presented in Table \ref{tab:performance_re}, we find that in the BC5CDR dataset, while LLaMA-2 achieves the highest recall, PaLM-2 performs the best in terms of Precision and F1. Meanwhile, in terms of F1, the zero-shot PaLM-2, Claude-2, and LLaMA-2 model even outperform the prior state-of-the-art fine-tuned BioGPT in this dataset, with an improvement of 17.61\% by the best performing PaLM-2. In the KD-DTI dataset, though GPT-3.5 and Claude-2 achieve high recall, their overall F1-score was quite lower than BioGPT and PaLM-2. Meanwhile, zero-shot PaLM-2 again performs much better while achieving almost similar performance in comparison to the fine-tuned BioGPT in terms of the F1 score. In the DDI dataset, GPT-3.5 achieves state-of-the-performance across all three metrics (Precision, Recall, and F1), followed by Claude-2. Since in the DDI dataset, there are only 4 types of labels, more descriptive prompts are used in this dataset (e.g., providing the definition of different interaction types), which helped GPT-3.5 and Claude-2 to achieve better performance. However, more descriptive prompts were not helpful for PaLM-2 in this dataset. Nonetheless, the impressive results achieved by LLMs in comparison to the prior state-of-the-art results in BC5CDR and DDI datasets demonstrate that in datasets having smaller training sets (both datasets have less than 1000 training samples), LLMs are more effective than even fine-tuned models. Meanwhile, in the KD-DTI dataset that has about 12K training samples, most zero-shot LLMs still achieve comparable performance, with PaLM-2 slightly outperforming the state-of-the-art result. More interestingly, while other LLMs achieve quite poor precision scores in the KD-DTI dataset, PaLM-2 even outperforms the current state-of-the-art result in terms of precision. {However, based on paired t-test with $p$ $\leq$ $.05$, the performance difference between the LLMs and the current fine-tuned SOTA models in terms of F1 is \textbf{not statistically significant}.}

\paragraph{\textbf{(ii) Text Classification:}}

In terms of Text Classification (see Table \ref{tab:performance_tc_qa_el}), {the LLM generated responses are evaluated based on \textbf{\textit{Hybrid Evaluation}}.} \textcolor{black} {In comparison to the current state-of-the-art models fine-tuned on the respective datasets (BioGPT \cite{luo2022biogpt} in HoC and BioBERT \cite{lee2020biobert} in LitCovid), it is evident that the zero-shot LLMs perform very poorly in comparison to the state-of-the-art fine-tuned baselines in both datasets}. In particular, the performance of Claude-2 was much poorer than other LLMs. Among LLMs, GPT-3.5 and PaLM-2 are generally better, with PaLM-2 being the best performing LLM in both the HoC dataset and the LitCovid dataset. 
{The difference in performance between the best performing PaLM-2 and the worst performing Claude-2 is also \textbf{statistically significant}, based on paired t-test, with $p$ $\leq$ $.05$}. 

We also investigate the effect of prompt tuning by evaluating two new prompts that are less descriptive, i.e., without giving definitions of the HoC classes, or without naming the HoC classes. Below our findings for GPT-3.5 based on prompt variations are demonstrated:

   
    \textit{(i) Prompting with only the name of each HoC class is given without any definitions, drops the F1 score to 46.93.}
    
    \textit{(ii) Prompting without explicitly mentioning the name of 10 HoC classes, drops F1 to 38.20.} 


 This indicates that for classification tasks, descriptive prompts are very helpful in improving the performance of LLMs (see Section \ref{sec:appendix_prompt_variation} for more details). 

\begin{figure*}
	\centering
		\includegraphics[height=4cm, width=16cm]{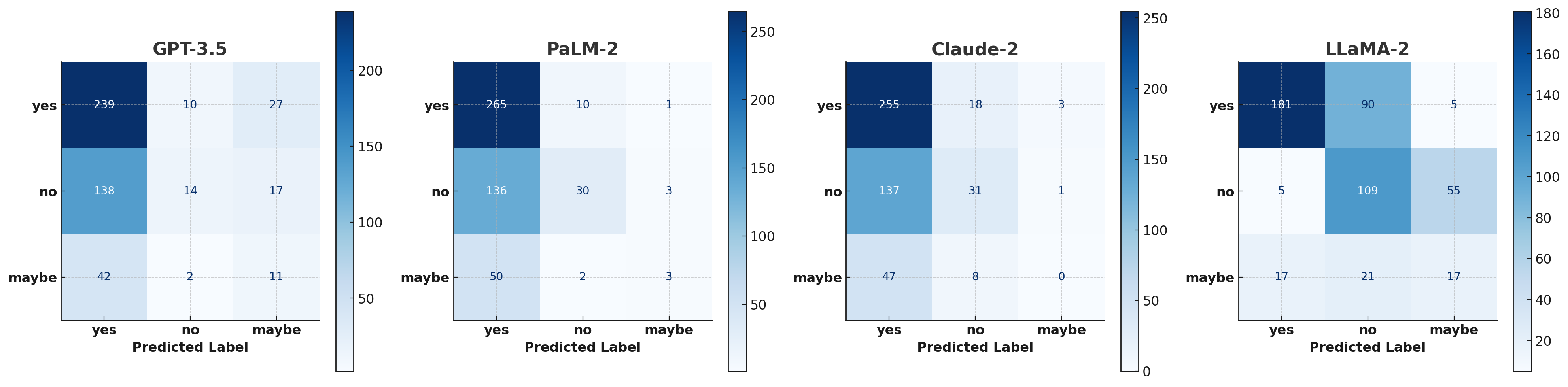}
	\caption{Confusion Matrix for different models in the PubMedQA dataset.}
	\label{fig:confusion_matrix}
\end{figure*}

\begin{figure*}
	\centering
		\includegraphics[height=4cm, width=16cm]{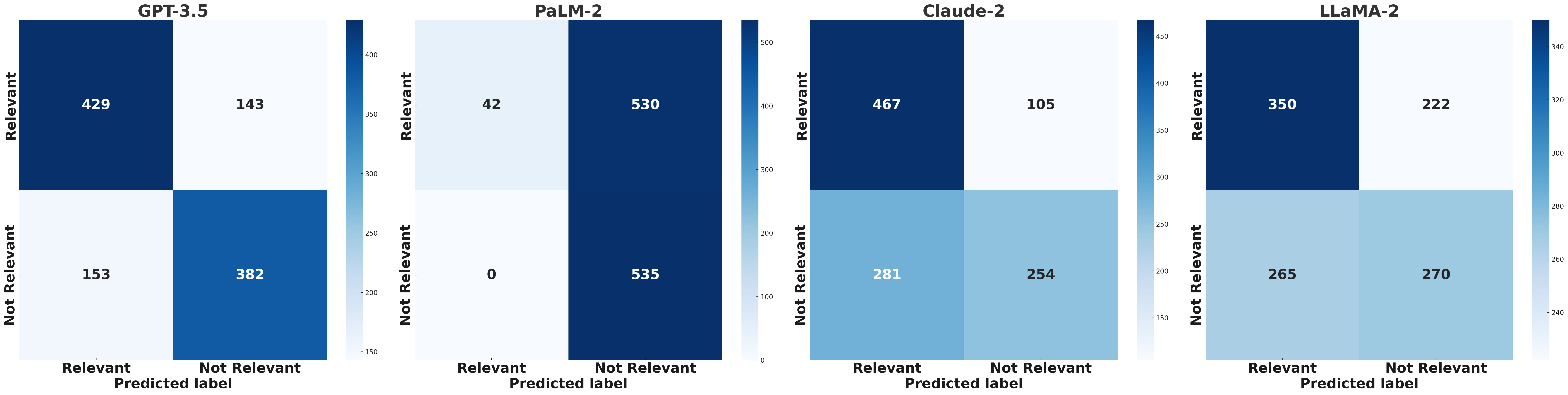}
	\caption{{Confusion Matrix for different models in the MediQA-2019 dataset.}}
	\label{fig:confusion_matrix_new}
\end{figure*}

\paragraph{\textbf{(iii) Question Answering:}}

{For question answering, we evaluate the performance based on \textbf{\textit{Hybrid Evaluation}} on two datasets  (see Table \ref{tab:performance_tc_qa_el})}. 

In terms of the question-answering task in the PubMedQA dataset, we find that the performance of all LLMs is much lower than the current state-of-the-art BioGPT model. {It should be noted that the BioGPT \cite{luo2022biogpt} model which achieves the state-of-the-art result in PubmedQA was additionally trained on the PQA-A (211K instances) and PQA-U (61K instances) splits of the PubmedQA dataset (along with the PQA-L split which is the dedicated training set of this dataset). While comparing the performance of the closed-source LLMs (GPT-3.5, PaLM-2, Claude-2), we find that they perform almost similarly, with none of them achieving more than 60\% accuracy.} More interestingly, none of these closed-source LLMs could outperform the LLaMA-2 model that achieves the best performance among LLMs in this dataset. This is an interesting finding since the LLaMA-2 only has 13B parameters, which is much smaller than the closed-source LLMs. To further investigate how LLaMA-2 achieves superior performance in this dataset, we present the confusion matrix using a heatmap based on the prediction made by different LLMs in Figure \ref{fig:confusion_matrix}. From the heatmap, we find that all LLMs except LLaMA-2 make mistakes while predicting the ``no'' type label, as in most cases the LLMs (GPT-3.5, PaLM-2, Claude-2) ended up predicted with the ``yes'' type label instead, leading to an overall poor accuracy.  

In terms of the question-answering task in the MediQA-2019 dataset, we find that the accuracy from the PubMedQA dataset is increased for GPT-3.5 and Claude-2, while being decreased for the LLaMA-2 and PaLM-2; with the zero-shot GPT-3.5 achieving the best accuracy (73.26). {The performance of GPT-3.5 is comparable to the current state-of-the-art accuracy of 79.49  \cite{he-etal-2020-infusing-mediqa2019} by the ALBERT model \cite{albert} which was additionally trained in question-answering style on 14K biomedical texts consisting of disease-related knowledge followed by being fine-tuned on the MediQA-2019 dataset. To further investigate the performance of LLMs in this dataset, we show the confusion matrix in Figure \ref{fig:confusion_matrix_new} to find that the best performing LLM in the MediQA-2019 dataset, the GPT-3.5 model 
was able to classify the Relevant and Not Relevant labels more accurately
than other LLMs. Moreover, the reason behind PaLM-2 being the worst performer in this dataset is due to the fact that it predicts most instances as Not Relevant. Paired t-test with $p$ $\leq$ $.05$ demonstrates that the performance difference between the LLMs in question answering is \textbf{not statistically significant.}}

\paragraph{\textbf{(iv) Entity Linking:}}

All the entity linking datasets are evaluated based on the \textbf{\textit{Hybrid Evaluation}} technique. For entity linking, we find from Table \ref{tab:performance_tc_qa_el} that Claude-2 outperforms all other LLMs in all three entity linking datasets: BC5CDR, Cometa, and NCBI. In BC5CDR and NCBI, \textcolor{black} {while LLaMA-2 is the second best performing model; the PaLM-2 is found to be the second best performer in the Cometa dataset}. Nonetheless, the performance of the second best performing models is still quite below in comparison to the Claude-2 model. This finding suggests that Claude-2 is more useful than other models in biomedical entity linking tasks by effectively retrieving the correct definition from its pre-training knowledge, although its performance is still much below compared to the current fine-tuned SOTA models, {which is also \textbf{statistically significant}, based on paired t-test with $p$ $\leq$ $.05$.}

\begin{table*}[t!]
\small
\setlength{\tabcolsep}{2.5pt} 
\caption{Performance on Named Entity Recognition datasets. SOTA results are from the BioBERT \cite{lee2020biobert} model. Here, `Precision' and  `Recall' are denoted by `P' and `R', respectively.}
\centering
\begin{tabular}{@{}c|ccc|ccc|ccc|ccc|ccc@{}}
\toprule
 & \multicolumn{15}{c}{\textbf{Model}   } \\ \cmidrule{2-16} 
\textbf{Dataset}  & \multicolumn{3}{c}{\textbf{GPT-3.5}} & \multicolumn{3}{c}{\textbf{PaLM-2}} & \multicolumn{3}{c}{\textbf{Claude-2}} & \multicolumn{3}{c}{\textbf{LLaMA-2-13b}} & \multicolumn{3}{c}{\textbf{SOTA}} \\ 
\cmidrule(lr){2-4} \cmidrule(lr){5-7} \cmidrule(lr){8-10} \cmidrule(lr){11-13} \cmidrule(lr){14-16}
&  \textbf{P} & \textbf{R} & \textbf{F1} 
&  \textbf{P} & \textbf{R} & \textbf{F1} 
&  \textbf{P} & \textbf{R} & \textbf{F1} 
&  \textbf{P} & \textbf{R} & \textbf{F1}
&  \textbf{P} & \textbf{R} & \textbf{F1}
\\ \midrule
\textbf{BC2GM}  
& 23.07 & 52.19 & 31.99 & 24.65 & 48.77 & 32.75 
& \textbf{31.95} & \textbf{55.10} & \textbf{40.45}  
& 3.39 & 24.11 & 5.95
& \textbf{84.32} & \textbf{85.12} & \textbf{84.72}
\\
\textbf{BC4CHEMD} 
& 17.33 & 52.08 & 26.01 & 18.27 & 44.09 & 25.83  
& \textbf{26.37} & \textbf{52.83} & \textbf{35.18}
& 3.67 & 35.05 & 6.64
& \textbf{92.80} & \textbf{91.92} & \textbf{92.36}

\\
\textbf{BC5CDR-chem}  
& {29.93} & {66.30} & {41.25} & {37.93} & {65.63} & {48.08} 
& \textbf{49.99} & \textbf{69.23} & \textbf{58.05}
& 6.98 & 48.41 & 12.21
& \textbf{93.68} & \textbf{93.26} & \textbf{93.47}
 \\ 
\textbf{BC5CDR-disease}  
& 23.37 & 52.08 & 32.26 & 26.56 & 46.16 & 33.72 
& \textbf{47.06} & \textbf{53.62} & \textbf{50.13} 
& 3.16 & 27.98 & 5.68
& \textbf{86.47}  & \textbf{87.84} & \textbf{87.15} \\ 

\textbf{JNLPBA} 
& 23.51 & 49.53 & 31.89 & 15.43 & 33.74 & 21.18 
& \textbf{26.97} & \textbf{48.34} & \textbf{34.62} 
& 2.50 & 15.32 & 4.30
& \textbf{72.24} & \textbf{83.56} & \textbf{77.49}
\\
\textbf{NCBI-disease} 
& 24.76 & 51.25 & 33.39 & 25.10 & 41.04 & 31.15  
& \textbf{39.33} & \textbf{54.69} & \textbf{45.75} 
& 2.56 & 21.67 & 4.58
& \textbf{88.22} & \textbf{91.25} & \textbf{89.71}
\\
\textbf{linnaeus}  
& 2.87 & 24.84 & 5.14 & 3.81 & 20.80 & 6.44 
& \textbf{8.30} & \textbf{42.92} & \textbf{13.91} 
& 0.73 & 24.21 & 1.42
& \textbf{90.77} & \textbf{85.83} & \textbf{88.24}
\\
\textbf{s800} 
& 9.38 & 45.89 & 15.57 & 10.80 & 39.50 & 16.96  
& \textbf{15.74} & \textbf{51.11} & \textbf{24.07} 
& 0.99 & 17.21 & 1.87
& \textbf{72.80} & \textbf{75.36} & \textbf{74.06}
\\ \bottomrule
\end{tabular}

\label{tab:result_ner}
\end{table*}

\paragraph{\textbf{(v) NER:}}
{Similar to Entity Linking, we also conduct \textbf{\textit{Hybrid Evaluation}} for NER and find from Table \ref{tab:result_ner} that Claude-2 again outperforms the rest other LLMs across all NER datasets (also in terms of all evaluation metrics: \textit{Precision}, \textit{Recall}, and \textit{F1}). However, the performance of all LLMs is significantly lower than the current SOTA results (based on paired t-test, this difference in performance is \textbf{statistically significant}, with $p$ $\leq$ $.05$), with the performance of LLaMA-2 being the poorest}. Such limitations of zero-shot LLMs in NER have also been observed in datasets from the general NLP domain \cite{laskar-etal-2023-systematic}. These findings give a strong indication that generative LLMs need further improvement on sequence labeling tasks like NER using the traditional BIO formatting. 

\paragraph{\textbf{(vi) Summarization:}}

\begin{table*}[t!]
\small
\centering
\setlength{\tabcolsep}{0.8pt} 
\caption{\small{Performance on various summarization datasets. Here, `R-1', `R-2', `R-L' and `B-S' denote `ROUGE-1', `ROUGE-2', `ROUGE-L', and `BERTScore', respectively.
State-of-the-art (SOTA) results are taken from the BioBART \cite{yuan2022biobart} model. Also, LLaMA-2 refers to its 13b version, similar to other tasks.}}

\begin{tabular}{c|cccc|cccc|cccc|cccc|cccc|ccccc}
\toprule
& \multicolumn{24}{c}{\textbf{Dataset}}  \\ \cmidrule(lr){2-25} 
{\textbf{Model}}  & \multicolumn{4}{c}{\textbf{iCliniq}} & \multicolumn{4}{c}{\textbf{HealthCareMagic}}  & \multicolumn{4}{c}{\textbf{MeQSum}}  & \multicolumn{4}{c}{\textbf{MEDIQA-QS}} & \multicolumn{4}{c}{\textbf{MEDIQA-MAS}} & \multicolumn{4}{c}{\textbf{MEDIQA-ANS}}  \\ 
\cmidrule(lr){2-5}\cmidrule(lr){6-9}\cmidrule(lr){10-13}
\cmidrule(lr){14-17}\cmidrule(lr){18-21} \cmidrule(lr){22-25}
 & \textbf{R-1} & \textbf{R-2} & \textbf{R-L} & \textbf{B-S} & \textbf{R-1} & \textbf{R-2} & \textbf{R-L} & \textbf{B-S} & \textbf{R-1} & \textbf{R-2} & \textbf{R-L} & \textbf{B-S} & \textbf{R-1} & \textbf{R-2} & \textbf{R-L} & \textbf{B-S} & \textbf{R-1} & \textbf{R-2} & \textbf{R-L} & \textbf{B-S} & \textbf{R-1} & \textbf{R-2} & \textbf{R-L} & \textbf{B-S} \\ 
\midrule

\textbf{GPT-3.5} 
& \textbf{30.5} & \textbf{12.8} & \textbf{25.4} & \textbf{89.3}
& \textbf{28.1} & \textbf{9.8} & \textbf{24.0} & \textbf{88.9}
& 30.0 & 12.3 & 26.2  & 89.0
& 30.6 & 11.6 & 26.7  & 89.0
& \textbf{38.9} & \textbf{14.6} & \textbf{22.1} & \textbf{87.9}
& \textbf{28.7} & \textbf{10.4} & \textbf{24.4}  & \textbf{89.0}
 \\ 
 \textbf{PaLM-2}
 & 21.9 & 10.2 & 18.6 & 87.0 & 25.9 & 9.8 & 22.0 & 88.3 & 31.5 & 14.0 & 27.7 & 89.7 & 29.7 & 11.5 & 26.0 & 90.0 & 15.3 & 8.6 & 13.5 & 85.2 & 25.4 & 12.1 & 18.9 & 85.4 \\

\textbf{Claude-2} 
 & 28.8 & 11.0 & 23.7 & 89.0
& 24.4 & 7.4 & 20.3 & 88.2 
& \textbf{31.7} & \textbf{13.6} & \textbf{27.9} & \textbf{89.9}
& \textbf{32.0} & \textbf{13.5} & \textbf{27.7} & 90.2
& 13.4 & 6.2 & 11.1 & 85.6
& 28.6 & 8.7 & 17.6 & 85.9 \\

\textbf{LLaMA-2} 
 & 20.0 & 7.2 & 15.2 & 85.8
& 16.7 & 5.1 & 12.9 & 85.3 
& 21.2 & 7.3 & 17.1 & 85.5
& 23.3 & 8.6 & 17.7 & 86.2
& 13.7 & 11.2 & 13.2 & 86.6
& 28.0 & 9.6 & 17.4 & 85.3

 \\ 

 \midrule
 \textbf{SOTA} 
& \textbf{61.1} & \textbf{48.5} & \textbf{59.4}  & \textbf{94.1}
& \textbf{46.7} & \textbf{26.1} & \textbf{44.2}  & \textbf{91.9}
& \textbf{55.6} & \textbf{38.1} & \textbf{53.2}  & \textbf{93.3}
& \textbf{32.0} & {12.4} & \textbf{29.7}  & \textbf{90.3}
& 32.9 & 11.3 & 29.3  & 86.1
& 21.6 & 9.3 & 19.2  & 85.7
 \\ 
\bottomrule
\end{tabular}

\label{tab:performance_biobart}
\end{table*}

\begin{table*}[t!]
\small
\setlength{\tabcolsep}{3pt} 
\caption{Performance on the Biomedical Lay Summarization task. State-of-the-Art  results are from \citet{sotadevlaysummtask1}.}
\centering
\begin{tabular}{c|cccc|cccc}
\toprule
 & \multicolumn{8}{c}{\textbf{Dataset}}\\
 \cmidrule(r){2-9} 
\textbf{Model} & \multicolumn{4}{c}{\textbf{eLife}} & \multicolumn{4}{c}{\textbf{PLOS}}\\

 \cmidrule(r){2-5} \cmidrule(r){6-9} & \textbf{ROUGE-1} & \textbf{ROUGE-2} & \textbf{ROUGE-L} & \textbf{BERTScore} & \textbf{ROUGE-1} & \textbf{ROUGE-2} & \textbf{ROUGE-L} & \textbf{BERTScore}\\
\midrule

\textbf{GPT-3.5} & 33.88 & 8.64 & 17.15 & \textbf{84.49} & \textbf{41.11} & \textbf{11.41} & \textbf{21.74} & \textbf{86.11} \\

\textbf{PaLM-2} & 21.55 & 3.92 & 12.14 & 81.03 & 29.61 & 7.10 & 16.40 & 83.02 \\

\textbf{Claude-2} & \textbf{39.20} & \textbf{9.31} & \textbf{18.34} & 84.30 & 39.05 & 9.28 & 19.52 & 85.03 \\

\textbf{LLaMA-2-13b} & 38.53 & 8.69 & 18.10 & 83.18 & 38.58 & 11.15 & 20.14 & 84.69 \\

\midrule 
\textbf{State-of-the-Art}& \textbf{49.50} & \textbf{14.60} & \textbf{46.90} & \textbf{85.50} & \textbf{50.20} & \textbf{19.00} & \textbf{46.20} & \textbf{86.50}
 \\
\bottomrule
\end{tabular}

\label{tab:lay_summ}
\end{table*}

\begin{table*}[t!]

\small
\setlength{\tabcolsep}{3pt} 
\caption{Performance on Readability Controlled Summarization in the PLOS dataset. State-of-the-Art results are from \citet{sotadevlaysummtask2}.}
\centering
\begin{tabular}{c|cccc|cccc}
\toprule
 & \multicolumn{8}{c}{\textbf{Summarization Type}}\\
 \cmidrule(r){2-9} 
\textbf{Model} & \multicolumn{4}{c}{\textbf{Abstract}} & \multicolumn{4}{c}{\textbf{Lay Summarization}}\\
 \cmidrule(r){2-5} \cmidrule(r){6-9} 
 & \textbf{ROUGE-1 } & \textbf{ROUGE-2 } & \textbf{ROUGE-L } & \textbf{BERTScore} & \textbf{ROUGE-1 } & \textbf{ROUGE-2 } & \textbf{ROUGE-L } & \textbf{BERTScore }\\
\midrule

\textbf{GPT-3.5} & 39.65 & 11.01 & 20.76 & \textbf{85.64} & \textbf{39.13} & \textbf{9.57} & \textbf{20.00} & \textbf{85.63}\\

\textbf{PaLM-2} & 25.09 & 5.37 & 14.20 & 82.53 & 30.70 & 7.02 & 16.39 & 83.31\\

\textbf{Claude-2} & \textbf{42.25} & \textbf{13.05} & \textbf{21.53} & 85.46 & 36.16 & 7.82 & 17.68 & 84.47 \\

\textbf{LLaMA-2-13b} & 41.78 & {13.01} & 21.37 & 84.63 & 36.33 & 9.53 & 18.89 & 84.18 \\

\midrule

\textbf{State-of-the-Art} & \textbf{46.97} & \textbf{15.57} & \textbf{42.87} & 85.48 & \textbf{45.67} & \textbf{13.38} & \textbf{41.59} & {85.57} \\

\bottomrule
\end{tabular}

\label{tab:readibility_summ}
\end{table*}

\textcolor{black} {We present the results on the following summarization datasets}: \textit{Dialog Summarization}, \textit{Question Summarization}, and \textit{Answer Summarization} in Table \ref{tab:performance_biobart} and compare with BioBART \cite{yuan2022biobart}. {For evaluation \cite{laskar2022domain}, we use the following two \textbf{\textit{Automatic Evaluation}} metrics: (i) the widely used ROUGE \cite{rouge} metric, and (ii) the BERTScore \cite{zhang2019bertscore} metric.} For BERTScore, we use the RoBERTa-Large \cite{liu2019roberta} model for implementation. For all LLMs, the input context length of 2000 words has been used. 

We observe that in terms of the ROUGE metric, all LLMs perform much worse than BioBART in datasets that have dedicated training sets, such as iCliniq, HealthCareMagic, and MeQSum. Meanwhile, they perform on par with BioBART in the MEDIQA-QS dataset. \textcolor{black} {Among LLMs, in general, GPT-3.5 is found to be the best performer in these datasets.} More importantly, GPT-3.5 outperforms BioBART in both MEDIQA-ANS and MEDIQA-MAS datasets. Note that MEDIQA-ANS, MEDIQA-MAS, and MEDIQA-QS datasets do not have any dedicated training data and GPT-3.5 and other LLMs usually achieve comparable or even better performance in these datasets compared to the BioBART model fine-tuned on other related datasets \cite{yuan2022biobart}. This further confirms that zero-shot LLMs are more useful than domain-specific fine-tuned models in biomedical datasets that lack large training data.

We also present our findings on the biomedical lay summarization task in Table \ref{tab:lay_summ} and readability controlled summarization task in Table \ref{tab:readibility_summ}.

\begin{table*}[t!]
\small
\caption{Performance of different LLMs on Biomedical Lay Summarization datasets based on various input lengths.}
\setlength{\tabcolsep}{3pt} 
\centering
\begin{tabular}{c|c|cccc|cccc}
\toprule
\multicolumn{2}{c}{} &\multicolumn{8}{c}{\textbf{Dataset}}\\
 \cmidrule(r){3-10} 
\multicolumn{1}{c}{\textbf{Model}} & \multicolumn{1}{c}{\textbf{Length}} & \multicolumn{4}{c}{\textbf{eLife}} & \multicolumn{4}{c}{\textbf{PLOS}}\\
\cmidrule(r){1-1} \cmidrule(r){2-2} \cmidrule(r){3-6} \cmidrule(r){7-10}
\multicolumn{2}{c}{} &  \textbf{ROUGE-1} & \textbf{ROUGE-2} & \textbf{ROUGE-L} & \textbf{BERTScore} & \textbf{ROUGE-1} & \textbf{ROUGE-2} & \textbf{ROUGE-L} & \textbf{BERTScore}\\
\midrule
\textbf{GPT-3.5}& 2000  & {33.88} & 8.64 & 17.15 & \textbf{84.49} & {41.11} & 11.41 & 21.74 & 86.11 \\
\textbf{GPT-3.5}& 5000  & 33.62 & 8.77 & 17.21 & 84.45 & 41.41 & 11.65 & 21.89 & 86.17 \\
\textbf{GPT-3.5}& 10000  & 33.39 & 8.60 & 17.16 & 84.35 & \textbf{41.59} & \textbf{11.94} & \textbf{22.11} & \textbf{86.25} \\
\midrule
\textbf{PaLM-2} & 2000 & 21.55 & 3.92 & 12.14 & 81.03 & 29.61 & 7.10 & 16.40 & 83.02 \\
\textbf{PaLM-2}  & 5000 & 15.13 & 2.54 & 8.71 & 79.27 & 25.00 & 5.78 & 13.89 & 82.10 \\
\midrule

\textbf{Claude-2}  & 2000 & {39.20} & {9.31} & {18.34} & {84.30} & 39.05 & 9.28 & 19.52 & 85.03 \\
\textbf{Claude-2}  & 5000 & \textbf{39.43} & \textbf{9.42} & \textbf{18.38} & 84.20 & 38.79 & 9.09 & 19.26 & 84.92 \\
\textbf{Claude-2} & FULL & 38.97 & 9.09 & 18.05 & 83.95 & 39.16 & 9.31 & 19.30 & 84.85 \\
\bottomrule
\end{tabular}

\label{tab:lay_summ_len}
\end{table*}

\begin{table*}[t!]
\small
\setlength{\tabcolsep}{2pt} 
\caption{{Performance of different LLMs on Readability Controlled Summarization in the PLOS dataset based on various input lengths.}}
\centering
\begin{tabular}{c|c|cccc|cccc}
\toprule
\multicolumn{2}{c}{} &  \multicolumn{8}{c}{\textbf{Summarization Type}}\\
 \cmidrule(r){3-10} 
\multicolumn{1}{c}{\textbf{Model}} & \multicolumn{1}{c}{\textbf{Length}} & \multicolumn{4}{c}{\textbf{Abstract}} & \multicolumn{4}{c}{\textbf{Lay Summarization}}\\
\cmidrule(r){1-1} \cmidrule(r){2-2}  \cmidrule(r){3-6} \cmidrule(r){7-10} 
  
\multicolumn{2}{c}{} & \textbf{ROUGE-1 } & \textbf{ROUGE-2 } & \textbf{ROUGE-L } & \textbf{BERTScore} & \textbf{ROUGE-1 } & \textbf{ROUGE-2 } & \textbf{ROUGE-L } & \textbf{BERTScore }\\
\midrule
\textbf{GPT-3.5} & 2000 & 39.65 & 11.01 & 20.76 & 85.64 & 39.13 & 9.57 & 20.00 & 85.53\\
\textbf{GPT-3.5} & 5000 & 40.94 & 11.83 & 21.40 & 85.90 & 40.07 & 10.27 & 20.66 & 85.81\\

\textbf{GPT-3.5} & 10000 & 40.99 & 11.89 & 21.44 & \textbf{85.91} & \textbf{40.29} & \textbf{10.42} & \textbf{20.71} & \textbf{85.86}\\

\midrule 
\textbf{PaLM-2} & 2000 & 25.09 & 5.37 & 14.20 & 82.53 & 30.70 & 7.02 & 16.39 & 83.31\\
\textbf{PaLM-2} & 5000 & 21.98 & 4.63 & 12.38 & 81.55 & 25.05 & 5.43 & 13.81 & 82.03\\
\midrule

\textbf{Claude-2} & 2000 & 42.25 & 13.05 & 21.53 & 85.46 & 36.16 & 7.82 & 17.68 & 84.47 \\
\textbf{Claude-2} & 5000 & 43.27 & 13.60 & 22.29 & 85.67 & 37.97 & 8.58 & 18.56 & 84.66 \\

\textbf{Claude-2} & FULL & \textbf{43.89} & \textbf{13.88} & \textbf{22.49} & 85.72 & 38.97 & 9.09 & 18.05 & 83.95 \\

\bottomrule
\end{tabular}

\label{tab:readibility_summ_len}
\end{table*}

For the biomedical lay summarization task, we combine both abstract and article together and give as input to the models till the concatenated text reaches the maximum context length. For this task, we compare the performance of the LLMs in eLife and PLOS datasets. \textcolor{black} Based on the ROUGE scores, the Claude-2 model is found to be the best performing LLM in the eLife dataset with GPT-3.5 being the best-performing one in the PLOS dataset. However, none of the LLMs could outperform the current state-of-the-art in these datasets. While the performance of the LLMs is quite low in terms of ROUGE, they achieve much higher scores in terms of BERTScore, which is comparable to the state-of-the-art result. This shows a great discrepancy between the lexical matching based traditional ROUGE scoring and the contextual similarity-based BERTScore metric. 

The readability-controlled summarization task contains two sub-tasks: (i) abstract writing, and (ii) lay summary writing. Contrary to the previous task (i.e., biomedical lay summarization task), this time we only give an article as input without the abstract, as required by the task. We find that in writing the abstract of the given article, the Claude-2 model performs the best in terms of all ROUGE scores. However, in terms of BERTScore, GPT-3.5 slightly performs better than Claude-2. Interestingly, we find that in terms of the BERTScore, the GPT-3.5 model even outperforms the ROUGE-based SOTA models in both datasets. This further establishes the limitation of using ROUGE as a metric to evaluate LLMs for summarization \cite{laskar-etal-2023-systematic}. 

\textcolor{black} {Since the whole document cannot be given as input at once to these LLMs except Claude-2, we also investigate the performance using the following input context lengths (in terms of number of words); PaLM-2: 2000 and 5000, GPT-3.5: 2000, 5000, and 10000, and Claude-2: 2000, 5000, and full input document.} Since LLaMA-2 has a maximum context length of 4000 tokens (approximately 3000 words\footnote{\url{https://help.openai.com/en/articles/4936856-what-are-tokens-and-how-to-count-them}}), we exclude LLaMA-2 from this study. \textcolor{black} {The results for both tasks, biomedical lay summarization, and readability controlled summarization, can be found in Table \ref{tab:lay_summ_len} and Table \ref{tab:readibility_summ_len}, respectively}. Our experiments reveal that increasing the context length decreases the performance of PaLM-2 in both tasks across all datasets. Moreover, increasing the context length also does not help GPT-3.5 or Claude-2 to gain any substantial performance gain. This can be explained based on the work of Liu et al. \cite{liu2023lost}, where they find that LLMs tend to lose contextual information with the increase in sequence length, and especially they perform poorly in scenarios when they are required to generate responses based on utilizing the information that appears in the middle of the context. 

\textcolor{black} {The experimental results in these article summarization datasets demonstrate that using the context length of 2000 is good enough in terms of ROUGE and BERTScore metrics for both abstract and lay summarization. This context length should also be very helpful in terms of usage cost as well as time efficiency in comparison to using longer contexts \cite{laskar2023building}}

{Further performance analysis demonstrates that based on the paired t-test with $p$ $\leq$ $.05$, the performance difference in terms of the ROUGE score between all the LLMs and the current fine-tuned SOTA models in the summarization datasets \textbf{is statistically significant}, which also happens in terms of BERTScore for all LLMs except GPT-3.5.}

\subsection{Analysis}

In this section, we conduct further analysis on the performance of LLMs based on (i) variations in prompts, (ii) few-shot learning, and (iii) fine-tuning, alongside analyzing the performance of LLMs based on the (iv) possibility of data contamination. Below, the findings based on this analysis are demonstrated. 

\subsubsection{Effects of Prompt Variation}
\label{sec:appendix_prompt_variation}
The effects of prompt tuning in the HoC dataset have been investigated by evaluating the performance of GPT-3.5 based on the following prompt variations:

\begin{enumerate}[label=\roman*.]
\item Prompting with explicitly defining the 10 HoC classes achieves an F1 score of 59.26 (see Row 1 in Table \ref{tab:prompt_variation_chatgpt}).
    \item Prompting without mentioning the name of any HoC classes, drops F1 to 38.20 (see Row 2 in Table \ref{tab:prompt_variation_chatgpt}).
    \item Prompting with the name of each HoC class is given without providing the definition of each class, drops the F1 score to 46.93 (see Row 3 in Table \ref{tab:prompt_variation_chatgpt}).
\end{enumerate}

Thus, our findings demonstrate that more descriptive prompts yield better results. 

\begin{table*}[t!]
\centering
\small
\caption{{Effects of Prompt Variations in GPT-3.5 for the Document Classification Task in the HoC dataset.}}
\begin{tabular}{p{1cm}p{10cm}p{1cm}}
\toprule
\textbf{\#} & \textbf{Prompt} & \textbf{F1} \\
\midrule
1. & The 10 hallmarks of cancer taxonomy with their definitions are given below: \newline
(i) Sustaining proliferative signaling: Cancer cells can initiate and maintain continuous cell division by producing their own growth factors or by altering the sensitivity of receptors to growth factors. \newline
(ii) Evading growth suppressors: Cancer cells can bypass the normal cellular mechanisms that limit cell division and growth, such as the inactivation of tumor suppressor genes and/or insensitivity to antigrowth signals. \newline
(iii) Resisting cell death: Cancer cells develop resistance to apoptosis, the programmed cell death process, which allows them to survive and continue dividing. \newline
(iv) Enabling replicative immortality: Cancer cells can extend their ability to divide indefinitely by maintaining the length of telomeres, the protective end caps on chromosomes. \newline
(v) Inducing angiogenesis: Cancer cells stimulate the growth of new blood vessels, providing the necessary nutrients and oxygen to support their rapid growth. \newline
(vi) Activating invasion and metastasis: Cancer cells can invade surrounding tissues and migrate to distant sites in the body, forming secondary tumors called metastases. \newline
(vii) Deregulating cellular energetic metabolism: Cancer cells rewire their metabolism to support rapid cell division and growth, often relying more on glycolysis even in the presence of oxygen (a phenomenon known as the Warburg effect). \newline
(viii) Avoiding immune destruction: Cancer cells can avoid detection and elimination by the immune system through various mechanisms, such as downregulating cell surface markers or producing immunosuppressive signals. \newline
(ix) Tumor promoting inflammation: Chronic inflammation can promote the development and progression of cancer by supplying growth factors, survival signals, and other molecules that facilitate cancer cell proliferation and survival. \newline
(x) Genome instability and mutation: Cancer cells exhibit increased genomic instability, leading to a higher mutation rate, which in turn drives the initiation and progression of cancer. \newline 
Classify the following sentence in one of the above 10 hallmarks of cancer taxonomy. If cannot be classified, answer as "empty": \newline 
[SENTENCE]
& 59.26 \\

\midrule

2. & Is it possible to classify the following sentence in one of the 10 categories in the Hallmarks of Cancer taxonomy? If possible, write down the class. \newline
[SENTENCE] 
& 38.20 \\
\midrule
3. & Classify the sentence given below in one of the 10 categories (i. activating invasion and metastasis, ii. tumor promoting inflammation, iii. inducing angiogenesis, iv. evading growth suppressors, v. resisting cell death,vi. cellular energetics, vii. genomic instability and mutation, viii. sustaining proliferative signaling, ix. avoiding immune destruction, x. enabling replicative immortality) in the Hallmarks of Cancer taxonomy? If cannot be classified, answer as ``empty''. \newline
[SENTENCE] 
& 46.93 \\ 

\bottomrule
\end{tabular}

\label{tab:prompt_variation_chatgpt}
\end{table*}

\subsubsection{Effects of Few-Shot Learning}

{In the previous analysis, it has been found that variations in prompts, especially the utilization of more descriptive prompts could significantly impact the performance of LLMs in zero-shot scenarios. While the main focus of this work was to conduct zero-shot experiments using LLMs to address the lack of large annotated datasets in the biomedical domain, this section demonstrates the effect of the utilization of few-shot examples in the prompts. Since few-shot learning also leads to an increase in the
context length, which is a problem for LLMs that have limited context
length, in this paper, the Claude-2 model is selected for the few-shot experiments since it can consider significantly much longer contexts (100k tokens) than other
LLMs. Thus, using Claude-2 as the LLM for the few-shot learning experiments also helped us to address the context length issue. In the prompt, the few-shot examples are first included, followed by the task descriptions, as demonstrated in Section \ref{prompt_design}. The results from the few-shot experiments across all datasets are shown in Table \ref{tab:result_few_shot}. \\
Though few-shot learning usually leads to improvements in performance, in many tasks, few-shot learning is also found to be ineffective. For instance, Ye et al. \cite{ye2023comprehensive} demonstrated that in many language processing tasks, few-shot learning using LLMs achieves much poorer results in comparison to zero-shot learning. In our experiments, we also find that while few-shot learning is more effective than zero-shot in some tasks (e.g., better in terms of F1 in KD-DTI (1-shot) and BC5CDR (3-shot) for relation extraction\footnote{Few-shot learning leads to a decrease in performance in terms of Recall in comparison to zero-shot learning in all relation extraction dataests.}, in terms of Accuracy in MediQA-2019 (1-shot) and PubMedQA (3-shot) for question answering, as well as in some summarization datasets), the opposite happens in other tasks as well (e.g., NER, Entity Linking, etc.). Therefore our findings are consistent with Ye et al. \cite{ye2023comprehensive} to reveal that increasing few-shot examples from 0-shot to 1 or 3-shot does not necessarily improve the performance. }

{To further improve performance with few-shot, the task examples in few-shot prompts are required to be of high quality to ensure better performance while avoiding possible prediction biases towards the task examples. Thus, future work may investigate how to construct better examples for few-shot experiments with LLMs in the biomedical domain.}

 \subsubsection{Effects of Fine-Tuning}

{The few-shot learning experiment demonstrates that adding few-shot examples to the prompt does not lead to any performance gain in most biomedical tasks. Thus, in this section, we investigate whether the fine-tuning of LLMs could lead to performance gain. Since the main motivation of this paper is to investigate how LLMs could be used to address the lack of annotated datasets problem in the biomedical domain, only the datasets that have smaller training sets have been used for the fine-tuning experiment. This makes the fine-tuning experiment to be also consistent with the motivation of this paper which is to investigate the capability of LLMs in zero-shot scenarios in the biomedical domain to address the lack of large annotated dataset issue. For this reason, the PubMedQA dataset for question-answering (only 450 training samples), the MeQSum dataset (500 training samples) for summarization, the DDI (500 training samples), and the BC5CDR (664 training samples) datasets for relation extraction have been used for LLM fine-tuning. Nonetheless, many closed-source LLMs (e.g., PaLM-2, Claude-2) do not support fine-tuning, whereas fine-tuning GPT-3.5 significantly increases the cost during inference\footnote{\url{https://openai.com/blog/gpt-3-5-turbo-fine-tuning-and-api-updates}}. Thus, the fine-tuning experiment is conducted with a comparatively smaller open-source LLM: the LLaMA-2-7B-Chat\footnote{\url{https://huggingface.co/meta-llama/Llama-2-7b-chat-hf}} model and run for 3 epochs with the learning rate $2e-5$. These hyperparameters are selected since they lead to the best performance in the validation set. The results of the fine-tuning experiment are shown in Table \ref{tab:performance_fine_tuning}. From Table \ref{tab:performance_fine_tuning}, it is quite evident that fine-tuning is more useful than few-shot learning. In general, fine-tuning outperforms all the zero-shot and few-shot LLMs (except GPT-3.5 in the DDI dataset in terms of Recall and F1, even though the fine-tuned version achieves significantly better precision scores). Meanwhile, in the summarization dataset, the fine-tuned LLaMA-2-7B set a new state-of-the-art result. Moreover, it achieves almost similar performance in comparison to the state-of-the-art in the PubMedQA dataset for the question answering task (even though LLaMA-2-7B was only trained on 500 samples, the current state-of-the-art BioGPT \cite{luo2022biogpt} model was trained on 270K samples).}

\begin{table*}[t!]
\small
\setlength{\tabcolsep}{3pt} 
\caption{{Experimental Results for Few-Shot Learning. Here, `Readability-Controlled', `ROUGE', and `BERTScore' are denoted by `RC', `R', and `B-S', respectively.`}}
\centering
\begin{tabular}{@{}c|ccc|ccc|ccc|ccc@{}}
\toprule
\textbf{Dataset}  & \multicolumn{3}{c}{\textbf{Claude-2 (0-Shot)}} & \multicolumn{3}{c}{\textbf{Claude-2 (1-Shot)}} & \multicolumn{3}{c}{\textbf{Claude-2 (3-Shot)}}  & \multicolumn{3}{c}{\textbf{SOTA}} \\ 
 \cmidrule(lr){1-4} \cmidrule(lr){5-7} \cmidrule(lr){8-10} \cmidrule(lr){11-13} 
\textbf{NER} &  \textbf{Precision} & \textbf{Recall} & \textbf{F1} 
&  \textbf{Precision} & \textbf{Recall} & \textbf{F1} 
&  \textbf{Precision} & \textbf{Recall} & \textbf{F1}
&  \textbf{Precision} & \textbf{Recall} & \textbf{F1}
\\ \midrule
{BC2GM}  
& 31.95 & 55.10 & 40.45 & 29.88 & 51.89 & 37.92 
& {29.76} & {47.19} & {36.50}  
& \textbf{84.32} & \textbf{85.12} & \textbf{84.72}
\\
{BC4CHEMD} 
& 26.37 & 52.83 & 35.18 & 22.28 & 52.41 & 31.27  
& {26.87} & {51.12} & {35.23}

& \textbf{92.80} & \textbf{91.92} & \textbf{92.36}

\\
{BC5CDR-chem}  
& {49.99} & {69.23} & {58.05} & {46.27} & {59.07} & {51.89} 
&{49.27} & {65.61} &{56.28}

& \textbf{93.68} & \textbf{93.26} & \textbf{93.47}
 \\ 
{BC5CDR-disease}  
& 47.06 & 53.62 & 50.13 & 44.65& 52.71 & 48.35 
& {43.77} & {51.27} &{47.22} 
& \textbf{86.47}  & \textbf{87.84} & \textbf{87.15} \\ 

{JNLPBA} 
& 26.97 & 48.34 & 34.62 & 26.63 & 46.29 & 33.81 
& {27.38} & 44.11 & {33.79} 

& \textbf{72.24} & \textbf{83.56} & \textbf{77.49}
\\
{NCBI-disease} 
& {39.33} & {54.69} & {45.75} & 37.28 & 55.42 & 44.57  
& {35.69} & {49.48} & {41.47} 
& \textbf{88.22} & \textbf{91.25} & \textbf{89.71}
\\
{linnaeus}  
& {8.30} & {42.92} & {13.91} & 8.31 & 33.22 & 13.29 
& {14.43} & {40.13} & {21.23} 

& \textbf{90.77} & \textbf{85.83} & \textbf{88.24}
\\

{s800} 
& {15.74} & {51.11} & {24.07} & 19.54 & 49.54& 28.02  
& {15.45} & {47.59} & {23.32} 
& \textbf{72.80} & \textbf{75.36} & \textbf{74.06} \\

\midrule
\textbf{Relation Extraction} &  \textbf{Precision} & \textbf{Recall} & \textbf{F1} 
&  \textbf{Precision} & \textbf{Recall} & \textbf{F1} 
&  \textbf{Precision} & \textbf{Recall} & \textbf{F1}
&  \textbf{Precision} & \textbf{Recall} & \textbf{F1}
\\ \midrule
{BC5CDR}  
& 44.04 & \textbf{67.73} & 53.37 & \textbf{66.95} & 40.45 & 50.18 
& {62.17} & {{53.34}} & {\textbf{57.42}}  
& 49.52 & 43.25 & 46.17 
\\
{KD-DTI} 
& 17.99 & \textbf{72.73} & 28.84 & 39.43 & {55.32} & \textbf{46.04}  
& {36.80} & {13.93} & {20.21} 
& \textbf{40.00} & 39.72 & 38.42
\\
{DDI} 
& 39.27 & \textbf{46.60} & \textbf{42.62} & 30.69 & 28.80 & 29.72  
& {33.89} & {24.27} & {28.28} & {\textbf{41.70}} & {44.75} & {40.76} \\

\midrule
\textbf{Entity Linking} 
&  \multicolumn{3}{c|}{\textbf{Recall@1}} 
&  \multicolumn{3}{c|}{\textbf{Recall@1}}  
& \multicolumn{3}{c|}{\textbf{Recall@1}}
&  \multicolumn{3}{c}{\textbf{Recall@1}}
\\ \midrule
{BC5CDR}  
& \multicolumn{3}{c|}{78.01} 
& \multicolumn{3}{c|}{47.91}  
& \multicolumn{3}{c|}{55.68}   
& \multicolumn{3}{c}{\textbf{93.26}}  
\\
{Cometa} 
& \multicolumn{3}{c|}{53.29} 
& \multicolumn{3}{c|}{55.59}  
& \multicolumn{3}{c|}{56.99}   
& \multicolumn{3}{c}{\textbf{81.77}}  
\\
{NCBI} 
& \multicolumn{3}{c|}{70.21} 
& \multicolumn{3}{c|}{49.17}  
& \multicolumn{3}{c|}{47.60}   
& \multicolumn{3}{c}{\textbf{89.90}}  \\

\midrule
\textbf{Question Answering} 
&  \multicolumn{3}{c|}{\textbf{Accuracy}} 
&  \multicolumn{3}{c|}{\textbf{Accuracy}} 
&  \multicolumn{3}{c|}{\textbf{Accuracy}}
&  \multicolumn{3}{c}{\textbf{Accuracy}}
\\ \midrule
{PubMedQA}  
& \multicolumn{3}{c|}{57.20} 
& \multicolumn{3}{c|}{52.23}  
& \multicolumn{3}{c|}{62.80}   
& \multicolumn{3}{c}{\textbf{78.20}}  
\\
{MediQA-2019} 
& \multicolumn{3}{c|}{65.13} 
& \multicolumn{3}{c|}{68.65}  
& \multicolumn{3}{c|}{63.32}   
& \multicolumn{3}{c}{\textbf{79.49}}  
\\

\midrule
\textbf{Text Classification} &  \multicolumn{3}{c|}{\textbf{F1}}  
&  \multicolumn{3}{c|}{\textbf{F1}}  
&  \multicolumn{3}{c|}{\textbf{F1}} 
&  \multicolumn{3}{c}{\textbf{F1}}
\\ \midrule
{HoC}  
& \multicolumn{3}{c|}{34.93} 
& \multicolumn{3}{c|}{38.99}  
& \multicolumn{3}{c|}{43.78}   
& \multicolumn{3}{c}{\textbf{85.12}}
\\
{LitCovid} 
& \multicolumn{3}{c|}{7.60} 
& \multicolumn{3}{c|}{4.01}  
& \multicolumn{3}{c|}{6.27}   
& \multicolumn{3}{c}{\textbf{86.20}} \\

\midrule
\textbf{Summarization} 
&  \multicolumn{3}{c|}{\textbf{R-1/R-2/R-L/B-S}} 
&    \multicolumn{3}{c|}{\textbf{R-1/R-2/R-L/B-S}}
&    \multicolumn{3}{c|}{\textbf{R-1/R-2/R-L/B-S}}
&   \multicolumn{3}{c}{\textbf{R-1/R-2/R-L/B-S}}
\\ \midrule

{iCliniq}  
& \multicolumn{3}{c|}{28.8/11.0/23.7/89.0}
& \multicolumn{3}{c|}{30.9/12.4/25.9/88.9}
& \multicolumn{3}{c|}{29.8/11.4/24.2/88.8}
& \multicolumn{3}{c}{\textbf{61.1/48.5/59.4/94.1}}
\\

{HealthCareMagic} 
& \multicolumn{3}{c|}{24.4/7.4/20.3/88.2}
& \multicolumn{3}{c|}{24.9/7.2/20.4/87.7}
& \multicolumn{3}{c|}{24.9/7.9/20.6/87.9}
& \multicolumn{3}{c}{\textbf{46.7/26.1/44.2/91.9}}
\\

{MeQSum} 
& \multicolumn{3}{c|}{31.7/13.6/27.9/89.9}
& \multicolumn{3}{c|}{26.8/10.6/22.4/87.7}
& \multicolumn{3}{c|}{29.1/11.7/24.8/88.2}
& \multicolumn{3}{c}{\textbf{55.6/38.1/53.2/93.3}} \\

{MEDIQA-QS} 
& \multicolumn{3}{c|}{32.0/\textbf{13.5}/27.7/90.2}
& \multicolumn{3}{c|}{26.8/11.0/21.8/88.1}
& \multicolumn{3}{c|}{27.7/11.0/22.02/88.2}
& \multicolumn{3}{c}{\textbf{32.0}/12.4/\textbf{29.7/90.3}} \\

{MEDIQA-MAS} 
& \multicolumn{3}{c|}{{13.4/6.2/11.1/85.6}}
& \multicolumn{3}{c|}{\textbf{36.5/11.4/20.3/86.7}}
& \multicolumn{3}{c|}{36.3/11.4/20.3/86.7}
& \multicolumn{3}{c}{32.9/11.3/29.3/86.1} \\

{MEDIQA-ANS} 
& \multicolumn{3}{c|}{28.6/8.7/17.6/85.9}
& \multicolumn{3}{c|}{30.9/10.8/19.6/86.3}
& \multicolumn{3}{c|}{\textbf{31.5/11.8/20.7/86.5}}
& \multicolumn{3}{c}{21.6/9.3/19.2/85.7}\\

{eLife (Lay Summ)} 
& \multicolumn{3}{c|}{39.2/9.3/18.3/84.3}
& \multicolumn{3}{c|}{39.3/8.9/17.9/84.1}
& \multicolumn{3}{c|}{37.6/8.5/17.5/84.1}
& \multicolumn{3}{c}{\textbf{49.5/14.6/46.9/85.5}} \\

{PLOS (Lay Summ)} 
& \multicolumn{3}{c|}{39.1/9.3/19.5/85.0}
& \multicolumn{3}{c|}{38.7/8.8/18.8/84.8}
& \multicolumn{3}{c|}{38.8/8.83/18.9/84.9}
& \multicolumn{3}{c}{\textbf{50.2/19.0/46.2/86.5}} \\

{PLOS (RC: Abstract)} 
& \multicolumn{3}{c|}{42.3/13.1/21.5/85.5}
& \multicolumn{3}{c|}{42.4/12.8/21.5/85.4}
& \multicolumn{3}{c|}{42.7/12.7/21.5/85.5}
& \multicolumn{3}{c}{\textbf{47.0/15.6/42.9/85.5}} \\

{PLOS (RC: Lay Summ)} 
& \multicolumn{3}{c|}{36.2/7.8/17.7/84.5}
& \multicolumn{3}{c|}{38.0/8.2/18.3/84.6}
& \multicolumn{3}{c|}{37.1/7.7/17.8/84.5}
& \multicolumn{3}{c}{\textbf{45.7/13.4/41.6/85.6}} \\

\bottomrule

\end{tabular}

\label{tab:result_few_shot}
\end{table*}

 \begin{table*}[t!]
 \setlength{\tabcolsep}{4pt} 
\small
\centering
\caption{{Experimental Results for Fine-Tuning. Here, `ROUGE' and `BERTScore' are denoted by `R' and `B-S', respectively.}}
\begin{tabular}{c|cccccc|c|cccc}
\toprule

& \multicolumn{6}{c|}{\textbf{Relation Extraction Task}} 
& \multicolumn{1}{c|}{\textbf{QA Task}}
& \multicolumn{4}{c}{\textbf{Summarization Task}} \\  

\cmidrule(lr){2-12}

   \textbf{Model}  & \multicolumn{3}{c}{\textbf{BC5CDR}} & \multicolumn{3}{c|}{\textbf{DDI}} & \multicolumn{1}{c|}{\textbf{PubMedQA}}  & \multicolumn{4}{c}{\textbf{MeQSum}}  \\ 
\cmidrule(lr){2-4}\cmidrule(lr){5-7}\cmidrule(lr){8-12}

& \textbf{Precision} & \textbf{Recall} & \textbf{F1} & \textbf{Precision} & \textbf{Recall} & \textbf{F1}  & \textbf{Accuracy} 
& \textbf{R-1} & \textbf{R-1} & \textbf{R-L} & \textbf{B-S} \\ 
\midrule

\textbf{GPT-3.5 (0-Shot)} & {30.62} & {73.85} & {43.29} & {47.11} & {45.77} & \textbf{46.43} & 54.40 & 30.0& 12.3& 26.2& 89.0 \\ 

\textbf{PaLM-2 (0-Shot)} & {51.61} & {57.30} & {54.30}  & {35.47} & {16.48} & {22.50} & 59.60 & 31.5& 14.0& 27.7& 89.7 \\ 

\textbf{Claude-2 (0-Shot)} & {44.04} & {67.73} & {53.37} & {39.27} & \textbf{46.60} & {42.62} & 57.20 & 31.7 &13.6 &27.9& 89.9  \\ 

\textbf{LLaMA-2-13b (0-Shot)} & {39.54} & \textbf{81.66} & {53.28} & {22.58} & {25.67} & 24.03 & 61.40 & 21.2& 7.3& 17.1& 85.5   \\ 

\textbf{Claude-2 (1-Shot)} & 66.95 & 40.45 & 50.18  & 30.69 & 28.80 & 29.72 & 52.23 & 26.8 & 10.6 & 22.4 & 87.7  \\ 

\textbf{Claude-2 (3-Shot)} & 62.17 & 53.34 & 57.4 & 33.89 & 24.27 & 28.28 & 62.80 & 29.1 & 11.7 & 24.8 & 88.2   \\ 

\textbf{LLaMA-2-7b (Fine-Tuned)} & \textbf{69.28} & {49.86} & \textbf{57.99} & \textbf{60.57} & {32.15} & {42.00} & 78.00 & \textbf{55.8 }& \textbf{38.4} & \textbf{53.6} & \textbf{91.7}   \\ 

\midrule
\textbf{SOTA} & 49.52 & 43.25 & 46.17 & 41.70 & 44.75 & 40.76 & \textbf{78.20} & 55.6& 38.1& 53.2 &93.3 \\ 
\bottomrule

\end{tabular}

\label{tab:performance_fine_tuning}
\end{table*}

\subsubsection{Data Contamination Detection Analysis}

{We follow the work of Li et al. \cite{li2023taskcontamination} to analyze the possibility of the contamination of the datasets that we study in this paper to evaluate various LLMs. For this purpose, we do the following similar\footnote{We did not compare the performance of LLMs based on the chronological analysis (which was also used by Li et al. in \cite{li2023taskcontamination}) since most of the classification datasets that have been used in this paper came before the data cut-off date of different LLMs.} to their work \cite{li2023taskcontamination}.}

\begin{enumerate}[label=\roman*.]
    \item {\textbf{Task Example Extraction:} This contamination detection technique checks whether the task example of a particular dataset (evaluated on discriminative tasks, i.e., non-summarization) can be extracted from the LLMs that we evaluated in this paper.}
    \item {\textbf{Membership Inference:} This contamination detection technique checks whether the response generated by LLMs in a particular dataset (evaluated on generation tasks, i.e., summarization) is an exact match of any gold labels in that dataset.}
\end{enumerate}

\begin{table*}[t!]
\small
\setlength{\tabcolsep}{5pt} 
\caption{{Data Contamination Detection Analysis. Here, `Task Example Extraction' and `Membership Inference' are denoted by `TEE' and `MI', respectively; whereas `NO' indicates that the possibility of contamination is not found, and `YES' indicates that there is a possibility of contamination found.}}
\centering
\begin{tabular}{@{}c|c|c|c|c@{}}
\toprule

\textbf{Task \& Dataset}  & \textbf{GPT-3.5 }& \textbf{PaLM-2} & \textbf{Claude-2} & \textbf{LLaMA-2-13B} \\ 
\cmidrule(lr){1-1} \cmidrule(lr){2-2} \cmidrule(lr){3-3} \cmidrule(lr){4-4} \cmidrule(lr){5-5} 
\textbf{NER} &  
{\textbf{TEE}} 
&  {\textbf{TEE}} 
&  {\textbf{TEE}}
&  {\textbf{TEE}}
\\ \midrule
{BC2GM (2008)}  
& No & No & No & No  
\\
{BC4CHEMD (2016)} 
& No & No & No & No

\\
{BC5CDR-chem (2015)}  
& No & No & No & No

 \\ 
{BC5CDR-disease (2014)}  
& No & No & No & No
\\
{JNLPBA (2004)} 
& No & No & No & No
\\
{NCBI-disease (2016) } 
& No & No & No & No
\\
{linnaeus (2010)}  
& No & No & No & No
\\

{s800 (2013)} 
& No & No & No & No \\

\midrule

\textbf{Relation Extraction} &  
{\textbf{TEE}} 
&  {\textbf{TEE}}
&  {\textbf{TEE}}
&  {\textbf{TEE}} \\ \midrule

{BC5CDR (2016)}  
& No & No & No & No
\\
{KD-DTI (2022)} 
& No & \textbf{Yes} & No & No
\\
{DDI (2013)} 
& \textbf{Yes} & No & \textbf{Yes} & No
\\ \midrule
\textbf{Entity Linking} & {\textbf{TEE}} 
&  {\textbf{TEE}}
&  {\textbf{TEE}}
&  {\textbf{TEE}} \\ \midrule
{BC5CDR}  
& No & No & No & No
\\
{Cometa} 
& No & No & No & No
\\
{NCBI} 
& No & No & No & No 
\\

\midrule
\textbf{Question Answering} 
& {\textbf{TEE}} 
&  {\textbf{TEE}}
&  {\textbf{TEE}}
&  {\textbf{TEE}} 
\\ \midrule
{PubMedQA (2019)}  
& No & No & \textbf{Yes} & \textbf{Yes}
\\
{MediQA-2019 (2019) } 
& No & No & No & No
\\

\midrule
\textbf{Text Classification} 
& {\textbf{TEE}} 
&  {\textbf{TEE}}
&  {\textbf{TEE}}
&  {\textbf{TEE}} 
\\ \midrule
{HoC (2016)}  
& No & No & No & No
\\
{LitCovid (2020)} 
& No & \textbf{Yes} & No & No
\\

\midrule
\textbf{Summarization} 
& {\textbf{MI}} 
&  {\textbf{MI}}
&  {\textbf{MI}}
&  {\textbf{MI}} 
\\ \midrule

{iCliniq (2020)}  
& No & \textbf{Yes}  & \textbf{Yes}  & No
\\

{HealthCareMagic (2020)} 
& \textbf{Yes}  & \textbf{Yes}  & \textbf{Yes}  & \textbf{Yes} 
\\

{MeQSum (2019)} 
& \textbf{Yes}  & \textbf{Yes}  & \textbf{Yes}  & \textbf{Yes}  \\

{MEDIQA-QS (2021)} 
& No & No & No & No \\

{MEDIQA-ANS (2020)} 
& No & \textbf{Yes}  & No & No\\

{MEDIQA-MAS (2021)} 
& No & No & No & No \\

{eLife (Lay Summ) (2023)} 
& No & No & No & No \\

{PLOS (Lay Summ) (2023)} 
& No & No & No & No\\

{PLOS (RC: Abstract) (2023)} 
& No & No & No & No \\

{PLOS (RC: Lay Summ) (2023)} 
& No & No & No & No \\

\bottomrule

\end{tabular}

\label{tab:contamination}
\end{table*}

{The results of the data contamination detection analysis are shown in Table \ref{tab:contamination}. From Table \ref{tab:contamination}, it can be inferred that in the NER datasets, none of the LLMs could extract the task examples. This could be due to the fact that in our experiments, the LLMs were asked to determine the NER tag for each token based on the `BIO' format. Meanwhile, the LLMs could potentially be pre-trained differently for the NER task. In our analysis, we also find that while LLMs could explain the NER tasks, they cannot generate the task examples for each dataset in the expected `BIO format'. The experimental results demonstrate that the possible absence of the task examples in the pre-training data could probably be the reason behind LLMs performing very poorly in all NER datasets. A similar trend is also observed in the Entity Linking datasets where no possibility of data contamination is found based on the task extraction analysis technique.}

{However, in Relation Extraction, task examples could be extracted in the KD-DTI and the DDI datasets (while the task example extraction approach did not lead to the possibility of data contamination in BC5CDR). In the case of the KD-DTI dataset, the best-performing PaLM-2 model could extract task examples, whereas in the DDI dataset, two of the better-performing LLMs, GPT-3.5 and Claude-2, could also extract task examples. This may indicate that the possible presence of task examples in the LLM training data may be responsible for the improved performance of some LLMs in respective datasets.}  

{In terms of the question answering and the text classification datasets, the task example extraction techniques show no possibility of data contamination in MediQA-2019 and HoC datasets. This is quite surprising for GPT-3.5 in the MediQA-2019 dataset since it achieves performance comparable to the state-of-the-art. While for HoC, it is expected since all LLMs perform much poorer than the state-of-the-art. For the other question-answering and text classification datasets, LLaMA-2 and Claude-2 show the possibility of data contamination in the PubMedQA dataset. This may provide some explanations on why smaller LLaMA-2-13b outperforms other much larger LLMs in this dataset. In the LitCovid dataset, we only find that the PaLM-2 model has the possibility of data contamination (it also achieves the best result in comparison to other LLMs in this dataset).}

{In the summarization datasets, the contamination detection analysis is conducted based on the membership inference technique which demonstrates that PaLM-2 is more likely to generate some responses similar to the gold reference summaries, as it shows the possibility of membership inference-based contamination in the highest number of datasets (4 out of the 10 summarization datasets).  We also find that the HealthcareMagic and the MeQSum datasets are reported as contaminated based on membership inference for all four LLMs. However, in none of these datasets, LLMs could beat the state-of-the-art models (with the results being much lower in comparison to the reported state-of-the-art results). It should also be pointed out that the membership inference shows no possibility of contamination in datasets that are released in 2023.}

\section{Conclusions and Future Work}
In this paper, we evaluate LLMs in six benchmark benchmark biomedical tasks across 26 datasets. We observe that in datasets that have large training data, zero-shot LLMs usually fail to outperform the fine-tuned state-of-the-art models (e.g., BioBERT, BioGPT, BioBART, etc.). 
{However, they consistently outperform the fine-tuned baselines on tasks where the state-of-
the-art results were achieved based on fine-tuning
only on smaller training sets. While the LLMs that are studied in
this paper are massive language models with a billion of parameters, they are trained on diverse domains and so when evaluating their zero-shot capabilities, they usually fail to outperform various state-of-the-art biomedical task specific fine-tuned models. However, fine-tuning these LLMs even on smaller training sets significantly improves their performance. Thus, it could be useful to train biomedical domain-specific LLMs on biomedical corpora to achieve better performance in tasks related to the biological and the medicine domain.} Moreover, our findings demonstrate that the performance of these LLMs may vary across different datasets and tasks, as we did not observe a single LLM outperforming others across all datasets and tasks. Thus, our evaluation in this paper could give a good direction for future research as well as real-world usage while utilizing these LLMs to build task-specific biomedical systems. We also demonstrate that LLMs are sensitive to prompts, as variations in prompts led to a noticeable difference in results. Thus, we believe that our evaluation will help future research while constructing the prompts for LLMs for various tasks. 

In the future, we will extend our work to investigate the performance of LLMs on more biomedical tasks \cite{wang2021pre}, such as medical code assignment \cite{ji2021does}, drug design \cite{monteiro2023fsm}, healthcare \cite{clinicalbert}, 
protein sequence \cite{shah2021gt}, as well as on {low-resource languages \cite{phan2023enrichinglowresource}} and problems in information retrieval that require open-domain knowledge \cite{huang2005york,huang2009bayesian, yin2010survival}. We will also explore the ethical implications (e.g., privacy concerns \cite{khalid2023privacy}) of using LLMs in the biomedical domain. \textcolor{black}{Moreover, we will extend our work to study the multi-modal LLMs \cite{team2023gemini,chen2023gpt4v,zhang2023mllm,zhang2023pmc,moor2023medflamingo} in the biomedical image processing tasks alongside also studying whether fine-tuning smaller open-source LLMs \cite{fu2024tiny} could outperform existing fine-tuned state-of-the-art models in the biomedical domain.}

\section*{Acknowledgement}
We would like to thank the handling editor and all the five reviewers of the Computers in Biology and Medicine journal for their excellent review comments. This research is supported by the research grant (RGPIN-2020-07157) from the Natural Sciences and Engineering Research Council (NSERC) of Canada, the York Research Chairs (YRC) program, and the Generic (Minor/Startup/Other) research fund of York University. We also acknowledge Compute Canada for providing us with the computing resources to conduct experiments, as well as Anthropic for providing us early access to the Claude-2 API.


\bibliography{anthology,custom}
\bibliographystyle{acl_natbib}
\end{document}